\pdfoutput=1

\documentclass[11pt]{article}

\usepackage{acl}

\usepackage{times}
\usepackage{latexsym}

\usepackage[T1]{fontenc}

\usepackage[utf8]{inputenc}

\usepackage[utf8]{inputenc} 
\usepackage[T1]{fontenc}    
\usepackage{hyperref}       
\usepackage{url}            
\usepackage{booktabs}       
\usepackage{amsfonts}       
\usepackage{nicefrac}       
\usepackage{microtype}      
\usepackage{xcolor}         

\usepackage{float}
\usepackage{wrapfig}
\usepackage{lipsum}
\usepackage{courier}
\usepackage{amsmath,amsfonts,amssymb,amsthm}
\usepackage{multirow}
\usepackage{arydshln}
\usepackage{listings}
\usepackage{paralist}

\usepackage{xcolor}

\definecolor{codegreen}{rgb}{0,0.6,0}
\definecolor{codegray}{rgb}{0.5,0.5,0.5}
\definecolor{codepurple}{rgb}{0.65,0,0.95}
\definecolor{backcolour}{rgb}{0.95,0.95,0.92}

\lstdefinestyle{mystyle}{
    commentstyle=\color{codegreen},
    keywordstyle=\color{magenta},
    numberstyle=\tiny\color{codegray},
    stringstyle=\color{codepurple},
    basicstyle=\ttfamily\footnotesize,
    breakatwhitespace=false,         
    breaklines=true,                 
    captionpos=b,                    
    keepspaces=true,                                 
    numbersep=5pt,                  
    showspaces=false,                
    showstringspaces=false,
    showtabs=false,                  
    tabsize=2
}

\usepackage[export]{adjustbox}
\usepackage[title,toc,titletoc,page]{appendix}

\usepackage[utf8]{inputenc} 
\usepackage[T1]{fontenc}    
\usepackage{hyperref}       
\usepackage{url}            
\usepackage{booktabs}       
\usepackage{amsfonts}       
\usepackage{nicefrac}       
\usepackage{microtype}      
\usepackage{xcolor}         
\usepackage{subcaption}
\usepackage{adjustbox}

\usepackage{microtype}

\usepackage{inconsolata}

\title{ZYN: Zero-Shot Reward Models with Yes-No Questions for RLAIF}

\author{Víctor Gallego \\
  Komorebi AI \\
  \texttt{victor.gallego@komorebi.ai} }

\begin{document}
\maketitle
\begin{abstract}
In this work, we address the problem of directing the text generation of a language model (LM) towards a desired behavior, aligning the generated text with the preferences of the human operator. We propose using another, instruction-tuned language model as a critic reward model in a zero-shot way thanks to the prompt of a Yes-No question that represents the user preferences, without requiring further labeled data. This zero-shot reward model provides the learning signal to further fine-tune the base LM using Reinforcement Learning from AI Feedback (RLAIF); yet our approach is also compatible in other contexts such as quality-diversity search. Extensive evidence of the capabilities of the proposed ZYN framework is provided through experiments in different domains related to text generation, including detoxification; optimizing sentiment of movie reviews, or any other attribute; steering the opinion about a particular topic the model may have; and personalizing prompt generators for text-to-image tasks. Code is released at \url{https://github.com/vicgalle/zero-shot-reward-models}.
\end{abstract}

\section{Introduction}

Large language models (LLMs), trained on extensive text datasets, have demonstrated remarkable emergent capabilities in zero or few-shot learning within the natural language processing (NLP) domain \cite{Radford2019LanguageMA, brown2020language, openai2023gpt4}. However, these models often exhibit undesirable behaviors such as fabricating information, producing biased or harmful content, or failing to adhere to user instructions \cite{parrots, gehman2020realtoxicityprompts, weidinger2021ethical, kenton2021alignment, bommasani2021opportunities, tamkin2021understanding}. Yet the customization of these models to accommodate specific user preferences poses a significant challenge. Therefore, the mitigation of undesired behaviors and the enhancement of model adaptability to user preferences are highly sought-after attributes in language models. 

Reinforcement Learning from Human Feedback (RLHF) has recently demonstrated impressive outcomes in aligning large, pretrained language models with human preferences. By optimizing for key qualities such as harmlessness and helpfulness, this technique, as underscored by \cite{bai2022training}, has gained significant ground. Furthermore, this approach has proven its efficacy by attaining unprecedented results across a wide array of natural language tasks \cite{openai2023gpt4}. 

The conventional RLHF pipeline refines an initial, non-aligned LLM by employing an online RL algorithm like the popular Proximal Policy Optimization (PPO) \cite{schulman2017proximal}. The goal is to optimize the LLM to align accurately with human preferences. A key dependency of RLHF is the reward model, which is trained to predict the best alternative from a pair of two model outputs, $(o_1, o_2)$, generated from the same prompt $p$. However, the large-scale collection of human-ranked preference data, particularly of high quality, can be expensive and time-consuming. In order to mitigate this challenge, several alternatives such as Reinforcement Learning from AI Feedback (RLAIF) have been suggested, where human annotation is not a requisite for the availability of labels. RLAIF methods \cite{bai2022constitutional} aim to mimic human binary preferences by assigning scores to outputs $o_1$ and $o_2$ utilizing a LLM. It's worth noting that the LLM used for scoring frequently matches the one that initially generates the given outputs $(o_1, o_2)$. Unsurprisingly, these LLM-generated binary choices tend to be somewhat less accurate compared to actual human labels, yet they can be helpful in some contexts. Furthermore, since the same prompt $p$ is used to generate both outputs, these are of comparable quality, so having the LLM rank the two can be challenging.

In this work, we depart from preference models that act on pairwise preference data (choose the best between two alternatives) to design zero-shot reward models that score each generation independently in a zero-shot fashion, just by prompting the model with a yes-no question, or an ensemble of questions. The resulting framework, which we call ZYN, is a way of creating reward models without need for annotated data, leveraging an instruction-tuned model that will act as the critic that will guide the student language model. As such, the rewards that ZYN computes can be straightforwardly integrated into any RL-finetuning pipeline, such as PPO-based RLAIF. But ZYN can also be used within other contexts too, such as best-of-N sampling and quality-diversity search.

The structure of this paper is as follows. Section \ref{sec:related} gives a brief overview of related literature, focusing on RL and instruction fine-tuning of language models; Section \ref{sec:method} showcases our novel approach to design and use zero-shot reward models; Section \ref{sec:experiments} presents a wide battery of different experiments and results, confirming the general usefulness and applicability of our framework; and Section \ref{sec:conclusions} sums up with conclusions, limitations, and several avenues for further research.

\section{Background and Related Work}\label{sec:related}

\paragraph{Reinforcement Learning from Human Feedback (RLHF) and AI Feedback (RLAIF).} RLHF has been employed to fine-tune language models for text summarization and to create helpful and harmless chat assistants \cite{ziegler2019fine, bohm-etal-2019-better, stiennon2020learning, wu2021recursively}. A standard RLHF pipeline fine-tunes an initial unaligned LM using an RL algorithm such as PPO \cite{schulman2017proximal}, steering the LM to align with human preferences. The work of \cite{ouyang2022training} recently proposed a method for refining language models using a hybrid approach that combines supervised learning of user-instructed data with RLHF. Traditionally, RLHF has been employed with pairwise preference data, which employ user rankings to rank multiple generations \cite{bai2022training, ouyang2022training, glaese2022improving}. However, this approach can potentially limit the richness and diversity of the feedback due to sparse signal in the rewards. Utilizing natural language critiques for a generation is a promising alternative way of obtaining learning signal. Notably, \cite{saunders2022self} strove to enhance feedback to summaries through critiques. In this work, we delve into how critiques, framed as yes-no questions, serve as the way of prompting a reward model to improve the generations of a base language models.

RLAIF approaches (such as the Constitutional AI framework \cite{bai2022constitutional}) simulate human pairwise preferences by having a LM rank the alternative generations, instead of humans; typically the reward LM will be the same as the one used to generate the alternatives. Several recent techniques build upon the RLAIF principles: RLCD trains a preference model using simulated preference pairs that contain both a high-quality and low-quality example, generated using contrasting positive and negative prompts \cite{yang2023rlcd}; and CLAIF, which works by corrupting input sentences and then ask the LM to fill in the gaps to obtain sentence embeddings \cite{cheng-etal-2023-improving}. Our work diverges from previous approaches to RLAIF in that we use an instruction-tuned model as a reward in a zero-shot way, just by prompting it with yes-no questions that can give a scalar reward for each generation, without need to rank a pair of them simultaneously. 

\paragraph{Instruction Tuning with Model-Generated Data.} Instruction tuning is an emerging area that employs natural language instructions to initiate significant zero-shot performance on hitherto unseen tasks. When language models are fine-tuned with human-written instructions, evidence from instruction tuning confirms that they can adeptly adhere to generic language instructions \cite{weller-etal-2020-learning, wei2021finetuned, mishra2021cross, sanh2021multitask, wang2022benchmarking}. The work of \cite{wang2022self} recently demonstrated that instructions generated by the model itself can be used for instruction tuning, substantially augmenting the capacity of basic language models to respond to instructions. Extending from this, there are several studies that perform instruction tuning on pre-trained language models using model-generated instructions \cite{alpaca, vicuna2023, chen2023phoenix, gpt4all}. One of our contributions involves the strategic use of instruction-tuned models, such as Flan-T5 \cite{flan}, as zero-shot reward models.

Other works that explore the topic of self-refinement in LMs, but do not delve into RL nor instruction fine-tuning, are: self-refinement prompting \cite{madaan2023self}; GPT-automated evaluations \cite{liu2023gpteval}; Tree of Thoughts prompting \cite{yao2023tree}; SelfCheck for zero-shot checking reasoning chains \cite{miao2023selfcheck}; the DSP framework, which generates data from a prompt, and then uses that data towards a goal \cite{khattab2022demonstrate}; and zero-shot tool usage \cite{hsieh2023tool}.

\section{ZYN: Zero-Shot Reward Models with Yes-No Questions}\label{sec:method}

We now introduce our approach, Zero-Shot Reward Models with Yes-No Questions (ZYN), a novel framework to enable reward models from instruction-tuned models in a zero-shot way thanks to prompting with binary questions.

\subsection{Method}

To construct a zero-shot reward model, ZYN begins with an instruction-tuned model, such as Flan-T5 \cite{flan}. This model will act as the critic with respect the text outputs of the unaligned LM, computing a scalar reward $r$ according to desired attribute of the text. To do so, the critic model is prompted with two elements: $o$, the generated text to be evaluated; and $q$, an instruction in the form of a binary question that reflects the desired attribute, written in natural language. As an example, $o$ could be a movie review, and the instruction question $q$ could be {\color{blue}"Is this movie review positive?"}, in the case we would like to fine-tune the base LM towards positive movie reviews. With these elements, a zero-shot reward model can defined by implementing a function $r = f(o, q)$. Then, the computed rewards can be straightforwardly used within a RLAIF pipeline, such as PPO-based fine-tuning \cite{trlx}.

Several alternatives are now provided to define the previous reward model $f$. In all of them, the critic model is prompted with a template similar to this one: {\color{blue}\texttt{"Text:}$o$ \,\texttt{Question:}$q$ \texttt{Response:}"}. That is, the question is specified at the end of the prompt.

\paragraph{Logit of affirmative answer.} We compute the value $v_{Yes}(o, q)$ of the logit which represents the token \texttt{"Yes"} after prompting the critic model with the template with the pair $(o, q)$; and then simply define $f(o,q) = v_{Yes}(o, q)$, up to re-scaling of the reward. However, during preliminary experiments, we found that this led to early model collapse: optimizing for the \texttt{"Yes"} token also led to optimization of closely-related tokens such as \texttt{"No"}, so the intended guidance of the critic model was lost. This motivated the next formulation.

\paragraph{Contrasting affirmative and negative answers.} We also extract the value $v_{No}(o, q)$ of the logit \texttt{"No"}, which serves as a contrastive term when comparing the logit of the affirmative answer. We can define the reward as the probability $p_{Yes \succ No}$ of choosing the \texttt{"Yes"} answer versus \texttt{"No"} in the following way:
\begin{equation}\label{eq:bt}
\begin{split}
    r &= p_{Yes \succ No}(o,q) =\\ &=\dfrac{\exp \lbrace v_{Yes}(o, q) \rbrace }{\exp \lbrace v_{Yes}(o, q) \rbrace + \exp \lbrace v_{No}(o, q) \rbrace}.
\end{split}
\end{equation}
Note that the previous equation can be interpreted as a Bradley-Terry preference model \cite{bradley1952rank} in which we compare the preference of the critic model to the affirmative answer versus the negative one. This can be seen as a way of performing rank aggregation and obtain a more robust score (explain better).

Since RL fine-tuning is sensible to the scaling of the rewards, we experimented with several alternative derivations from Eq. (\ref{eq:bt}) that may improve stability during training. One alternative is to adopt the log-odds ratio to widen the range of rewards:
\begin{equation}\label{eq:logodds}
r = \log \dfrac{p_{Yes \succ No}(o,q)}{1 - p_{Yes \succ No}(o,q)}.
\end{equation}
Another variation is to re-scale Eq. (\ref{eq:bt}) to have rewards centered around 0 during PPO-training, as in
\begin{equation}\label{eq:bt_scaled}
    r = k_{s} \left( \dfrac{\exp \lbrace v_{Yes}(o, q) \rbrace }{\exp \lbrace v_{Yes}(o, q) \rbrace + \exp \lbrace v_{No}(o, q) \rbrace} - k_{c} \right),
\end{equation}
with $k_{s}, k_{c}$ being scaling and centering hiperparameters, respectively, that control the mean and the scale of the rewards. The choice between these alternatives for the reward function should be treated as a hyperparameter that should be selected for each task.
\paragraph{Ensemble of multiple questions.} Lastly, it is also possible to create an ensemble reward model given a set of $K$ critic questions, $\lbrace q_i \rbrace_{i=1}^K$, and average the rewards for each question with
\begin{equation}\label{eq:ensemble}
r = \sum_{i=1}^K w_i f(o, q_i)
\end{equation}
for any convex combination of $w_i$'s. By using more than one question prompt this also encourages getting more robust generations inside RL training, and avoids adversarial optimization to just one prompt, which can have negative results (i.e., reward hacking \cite{skalse2022defining}). \\

\begin{figure}[!ht]
\centering
\includegraphics[width=0.49\textwidth]{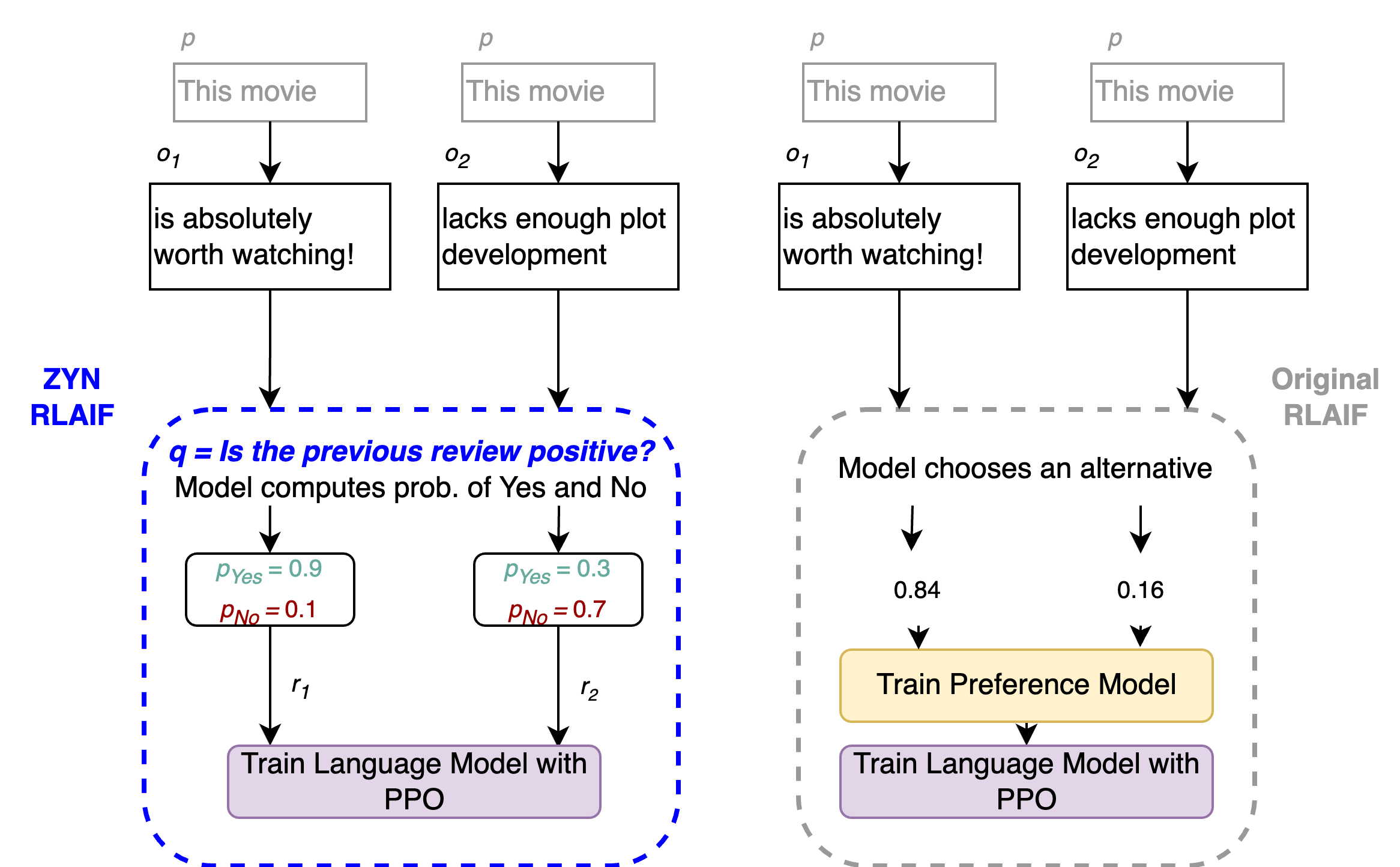}
\caption{Diagram depicting ZYN versus original RLAIF approaches, for an example of optimizing positive sentiment while generating movie reviews. The ZYN method scores each generation independently, without the need for specific reward model training because it leverages instruction-tuned models in a zero-shot manner with question prompting. Contrarily, vanilla RLAIF methods require the critic model to choose between two generations and the training of this particular preference model.
}\label{fig:zyn}
\end{figure}

Once a reward function $f(o,q)$ has been chosen, it can be integrated into any RL pipeline, such as PPO-based fine-tuning or any other training algorithm, to optimize the rewards by steering the output distribution of the base LM, as in RLHF. And it can also be used with other approaches that do not rely on RL, such as best-of-N sampling \cite{BoN} or quality-diversity search \cite{bradley2023qdaif}, in which we have the ZYN reward model choose the best generation. See Section \ref{sec:experiments} for examples of different applications. Figure \ref{fig:zyn} summarizes the fine-tuning process with the ZYN RLAIF framework.
\subsection{Implementation details}
ZYN is straightforward to implement, as it only requires access to the value of the tokens of the zero-shot, critic reward model. Thus, any instruction-tuned model can potentially be used with this framework, as it is agnostic of model architecture.
It is also computationally cheap, as it only requires inputting the prompt template and evaluating the corresponding affirmative and negative tokens, so no expensive auto-regressive sampling is necessary during the reward computation phase. Listing \ref{lst:code} in Appendix \ref{appendix:implementation} showcases a Python implementation of a ZYN reward model as a wrapper of a HuggingFace model.

\subsection{Comparison with related approaches}

\paragraph{Original RLAIF \cite{bai2022constitutional}.} Whereas vanilla RLAIF has a critic model choose the best generation between two alternative texts, the critic model from ZYN scores each generation independently, based on the probabilities of positive and negative answer to a Yes-No question. As such, ZYN doesn't require the specific training of a reward model, as it can leverage current instruction-tuned models in a zero-shot way. 

\paragraph{RLCD \cite{yang2023rlcd}.} In this approach, the underlying LM needs to be prompted with both a positive and negative prompt, which will serve as labels to train a custom pairwise preference model, which will act as the guide for the student LM. Similarly as before, our ZYN reward model does not score in a pairwise fashion, and we do not need to train an additional reward model: the prompting is transferred from the student LM to the critic model via yes-no questions.


\section{Experiments and Results}\label{sec:experiments}

In this Section, we aim to showcase the capabilities of the ZYN framework in a diverse set of tasks. The ease of use, its generality to different settings, and the positive results, are the main benefits of ZYN. We mainly explore two different settings: a battery of experiments in RLAIF (Section \ref{sec:rlaif}), and an example in quality-diversity search (Appendix \ref{sec:qd}).

\subsection{RLAIF}\label{sec:rlaif}

As the training procedure, for all experiments we use the PPO algorithm from the \texttt{trlx} library \cite{trlx}.

\subsubsection{Optimizing movie review sentiment}\label{sec:411}

The goal is to have a language model improve the sentiment (towards the positive side) of the movie reviews it generates. To this end, we experimented with two differently-sized language models as the students: a GPT-2 fine-tuned over the IMDB dataset of movie reviews (137M parameters); and GPT-Neo-1.3B \cite{gpt-neo}. As the critic reward model, we use Flan-T5-large \cite{flan}. As the question prompt, we use $q_1 =$ {\color{blue}"Is this movie review positive?"}. To avoid model collapse to degenerate outputs, we also use an additional question prompt $q_2 =$ {\color{blue}"Is this text too repetitive?"} (swapping the "\texttt{Yes}" and "\texttt{No}" labels in Eq (\ref{eq:bt})), and combine the two rewards with the ensemble formulation in (\ref{eq:ensemble}), using identical weights. As the evaluation set of prompts in the student LM, we sample 64 random reviews from IMDB, and take just the first two words.

\begin{figure}[!h]
\centering
\begin{subfigure}[b]{.49\textwidth}
  \centering
  \includegraphics[width=1\linewidth]{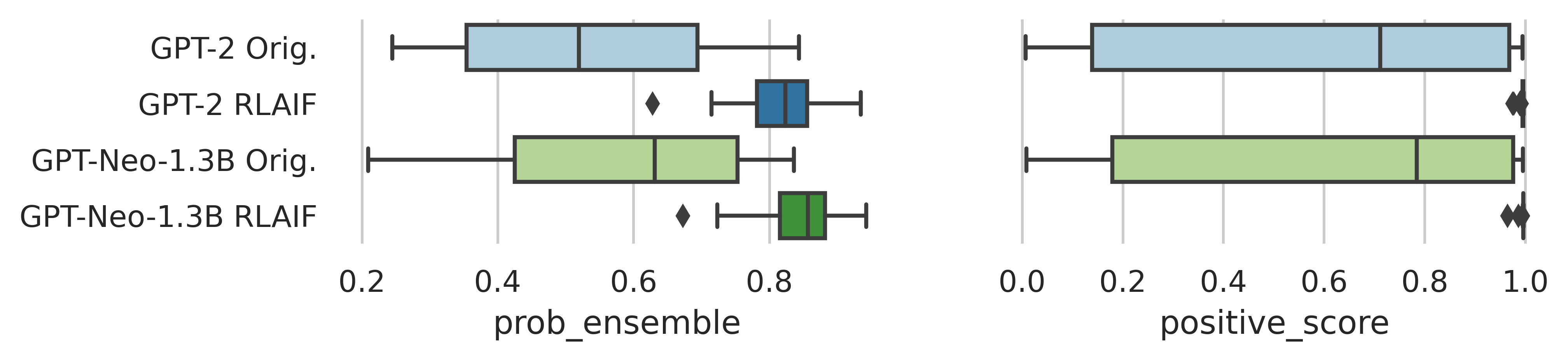}
  \caption{Sentiment distribution for the evaluation set of prompts}
  \label{fig:exp1_1}
\end{subfigure}%
\hspace{0.5cm}
\begin{subfigure}[b]{.46\textwidth}
\resizebox{\columnwidth}{!}{
 \begin{tabular}{rcccc}
\toprule
variation & reward & prob\_positive & prob\_ensemble & positive\_score \\
\midrule
GPT-2 Orig. & 0.34 & 0.51 & 0.53 & 0.59 \\
GPT-2 RLAIF & 3.20 & 0.97 & 0.82 & 0.99 \\
GPT-Neo-1.3B Orig. & 0.86 & 0.61 & 0.58 & 0.61 \\
GPT-Neo-1.3B RLAIF & 3.44 & 0.98 & 0.84 & 0.99 \\
\bottomrule \\ [0.1cm]
\end{tabular}
}
  \caption{Average reward and sentiment score for the evaluation set of prompts. The reward is computed using formula (\ref{eq:bt_scaled}) with $k_c = 0.5$ and $k_s = 10$}
  \label{fig:exp1_2}
\end{subfigure}
\caption{Experiment results for the optimizing movie review sentiment task}
\label{fig:exp1}
\end{figure}

Figure \ref{fig:exp1} shows the results, comparing the original models versus the ZYN RLAIF'ed ones. To evaluate the sentiment with a different approach than our zero-shot model, we used a fine-tuned classifier as the ground truth. The probability of positive sentiment assigned by this model is what we call \texttt{positive\_score}. Compare it with \texttt{prob\_ensemble}, which is the reward score of the ensemble of prompts computed via Eqs. (\ref{eq:bt}) and (\ref{eq:ensemble}). While for the original, unaligned LMs the sentiment scores were significantly wide, with both negative and positive values for the sentiment, with the help of ZYN RLAIF, the distribution of the generations is completely steered towards the positive side.


\subsubsection{Optimizing arbitrary movie review attributes}

Next, we tested whether ZYN can also guide the LM towards movie reviews that exhibit or focus on an arbitrary attribute specified by the user via the question prompt, in a zero-shot manner. We pick three question prompts that we deemed challenging yet diverse enough: $q_1 =$ {\color{blue}"Does this movie review focus on boring characters?"},
$q_2 =$ {\color{blue}"Does this movie review focus on an amazing and thrilling plot?"}, and
 $q_3 =$ {\color{blue}"Does this movie review sound professional?"}.

We use the same GPT-2 model as before as the student, and we test the ZYN RLAIF for each of the previous question prompts independently. Results are shown in Table \ref{tab:exp2}, demonstrating that ZYN has significant effect in optimizing the rewards of the model's generations towards the desired attributes. See Table \ref{tab:exp2_2} for sample generations for each of the attributes.

Critically, in this set of experiments we don't have ground truth labels. A recent trend in evaluating language models is to create automated evaluations using a bigger and more capable language model \cite{Chen2023ExploringTU}. As the evaluator, we use the \texttt{gpt-3.5-turbo} model. However, the API access doesn't provide logits of the response, so we resort to evaluating each generation using a Likert score that assesses how well the generation meets the question prompt. We compute the Spearman's $\rho$ correlation between the reward score given by the zero-shot reward model, and the rating score provided by \texttt{gpt-3.5-turbo}. Results per task are also in Table \ref{tab:exp2}. The Spearman's rho is significantly moderate to high in all cases, suggesting that our zero-shot reward model is a good proxy for steering the generated responses towards the desired attributes.
\begin{table}[!h]
\centering
\begin{adjustbox}{width=\columnwidth}
\begin{tabular}{rccc}
\toprule
Attribute & Config. & Reward Score & Spearman's $\rho$ \\
\midrule
\multirow[t]{2}{*}{boring characters} & Original LM & $0.22 \pm  0.14$ & \multirow[t]{2}{*}{$0.69$}\\
 & RLAIF'd LM & $\mathbf{0.53 \pm  0.24}$ \\
\cline{1-4} \\[-6pt]
\multirow[t]{2}{*}{an amazing thrilling plot} & Original LM & $0.42 \pm 0.28$ & \multirow[t]{2}{*}{$0.68$}\\
 & RLAIF'd LM & $\mathbf{0.60 \pm  0.27}$ \\
\cline{1-4} \\[-6pt]
\multirow[t]{2}{*}{sound professional} & Original LM & $ 0.58 \pm  0.22$ & \multirow[t]{2}{*}{$0.53$}  \\
 & RLAIF'd LM & $ \mathbf{0.80 \pm  0.18 }$ \\
\cline{1-4} \\[-6pt]
\multirow[t]{2}{*}{\textbf{Global Average}} & Original LM & $0.41 \pm  0.25 $ & \multirow[t]{2}{*}{$0.63$}\\
 & RLAIF'd LM & $\mathbf{0.64 \pm 0.24}$  \\
\bottomrule \\
\end{tabular}
\end{adjustbox}
\caption{Results for the optimizing arbitrary movie review attributes task. Reward Score is computed using Eq. (\ref{eq:bt}) for each of the question prompts, under the Flan-T5-large critic model. Spearman's $\rho$ correlates the Reward Score with a Likert score generated by \texttt{gpt-3.5-turbo}.}\label{tab:exp2}
\end{table}
\begin{table*}[!h]
\centering
\begin{tabular}{rllc}
\toprule
Attribute & Config. & Generation  & RS \\
\midrule
boring characters &
Original LM &
\multicolumn{1}{p{5cm}}{\raggedright \tiny {\textcolor{blue}{This is one of}} the most interesting parts of the film. He's not blending seamlessly in with the class-only crowd.<br /><br />1979 (Italy): Script editor Mario Castiglione (protocol writer}  & 0.20  \\
boring characters &
RLAIF'd LM &

\multicolumn{1}{p{5cm}}{\raggedright \tiny {\textcolor{blue}{This is one of}} Hollywood's worst 00s-style movies.<br /><br />The acting is awful, the script is ill-conceived and the script is a non-starter. The cast is bottomless }  & 0.75  \\
\midrule
an amazing thrilling plot & 
Original LM &

\multicolumn{1}{p{5cm}}{\raggedright \tiny {\textcolor{blue}{I thought this was}} going to be a great film. Watch at least a few of the crocodile deaths and you can almost see it. And..how could Synergizer want to break into Swedish and expected people to}  & 0.11 \\
an amazing thrilling plot & 
RLAIF'd LM &

\multicolumn{1}{p{5cm}}{\raggedright \tiny {\textcolor{blue}{I thought this was}} a great story. The storyline itself was excellent - the story is incredibly complex and suspenseful. I have seen it many times and you will see it every time. I enjoyed the story and series and}  & 0.98  \\

\midrule
sound professional
 & 
Original LM &

\multicolumn{1}{p{5cm}}{\raggedright \tiny {\textcolor{blue}{If you took a}} knee, you'd have left a bare knuckle in your forehead. But you must realize that for anyone to allow their teeth to lay flat with witnesses offstage and their lips exposed by \"an actor}  & 0.32  \\
sound professional &
RLAIF'd LM &

\multicolumn{1}{p{5cm}}{\raggedright \tiny {\textcolor{blue}{If you took a}} look at some of the other stories, I enjoyed them quite a bit. The film has quite a deal of skill development and is beautiful and well constructed and well done. And great performances certainly prepared the }  & 0.96  \\




\bottomrule
\end{tabular}
\vspace{0.2cm}
\caption{Sample generations for the optimizing movie review attributes task. RS is the reward score computed via (\ref{eq:bt})}\label{tab:exp2_2}
\end{table*}

\subsubsection{Detoxification}

We now turn into a detoxification task, in which we aim to reduce the amount of hate-speech or toxic content produced by a language model. As in \ref{sec:411}, we test ZYN RLAIF with two distantly-sized language models, GPT-Neo-1.3B and Pythia-410M \cite{pmlr-v202-biderman23a}. Note that neither of these two models are instruction-tuned, they are generalist pretrained language models. As the critic model we used Flan-T5-large, with the question prompt $q =$ {\color{blue}"Does this text contain toxic speech?"}.

\begin{figure}[!ht]
\centering
\begin{subfigure}[b]{.49\textwidth}
  \centering
  \includegraphics[width=1\linewidth]{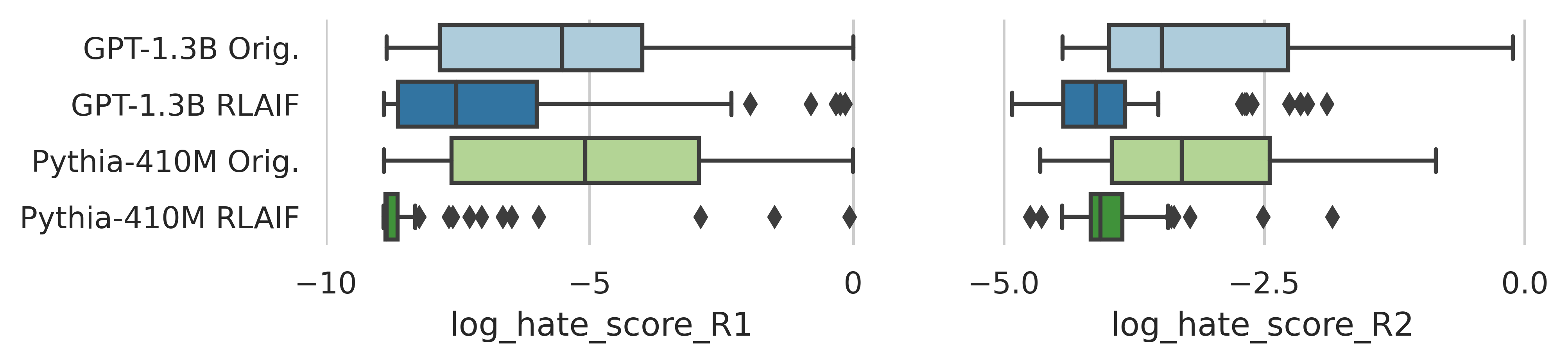}
  \caption{Hate-speech distribution (log space) for the evaluation set of prompts}
  \label{fig:exp3_11}
\end{subfigure}%
\hspace{0.5cm}
\begin{subfigure}[b]{.46\textwidth}
\resizebox{\columnwidth}{!}{
 \begin{tabular}{rcccccc}
\toprule
variation & reward & RS & HS\_R1 & HP\_R1 & HS\_R2 & HP\_R2 \\
\midrule
GPT-1.3B Orig. & -2.26 & 0.27 & 0.08 & 0.06 & 0.10 & 0.05 \\
GPT-1.3B RLAIF & -0.02 & 0.50 & 0.05 & 0.05 & 0.03 & 0.00 \\
Pythia-410M Orig. & -2.15 & 0.29 & 0.13 & 0.11 & 0.07 & 0.00 \\
Pythia-410M RLAIF & 2.45 & 0.74 & 0.02 & 0.02 & 0.02 & 0.00 \\
\bottomrule \\ [0.2cm]
\end{tabular}
}

  \caption{Average reward and hate-speech scores for the evaluation set of prompts}
  \label{fig:exp3_12}
\end{subfigure}
\caption{Results for the detoxification experiments}
\label{fig:exp3_1}
\end{figure}

Results are displayed in Figure \ref{fig:exp3_1}. As the ground truth, we evaluate the generations using two different hate-speech classifiers, \texttt{facebook/roberta-hate-speech-dynabench- r4-target} and \texttt{cardiffnlp/twitter-roberta- base-hate}, to validate our zero-shot reward model. The metric HS-R$i$ is the hate-speech score computed by each of the previous classifiers ($i=1,2$), and HP-R$i$ denotes the fraction of generations reported as hate-speech. RS is the zero-shot reward score computed with Eq. (\ref{eq:bt}) that guides the PPO training. Note that just by optimizing with this learning signal, that just requires the zero-shot prompt $q$, decreases the hate-speech of the generations, as reported both by the zero-shot model and the two independent classifiers.

\subsubsection{Opinions on gun ownership}

In this experiment the objective is to modify the opinion a LM has towards a topic, in particular gun ownership. As the student model we choose GPT-Neo-2.7B, and to elicit its opinion about the topic, we sourced a dataset of 20 questions related to gun ownership and usage from OpinionQA \cite{santurkar2023opinions}, which was split into training and evaluation sets (refer to Tables \ref{tab:gun1} and \ref{tab:gun2} in the Appendix to see the list of questions). During the PPO training, we have the model answer these questions, and then use the zero-shot reward model to score the answers, depending on whether they are pro or against guns. As the zero-shot reward model, we use Flan-T5-XL with the prompt question $q =$ {\color{blue}"Is the AI against guns?"}. We perform two independent training runs: one for optimizing the opinion towards being "against guns", and the other for being "pro guns", just by flipping the sign of the rewards.

\begin{figure}[!h]
\centering
\begin{subfigure}[b]{.49\textwidth}
  \centering
  \includegraphics[width=1\linewidth]{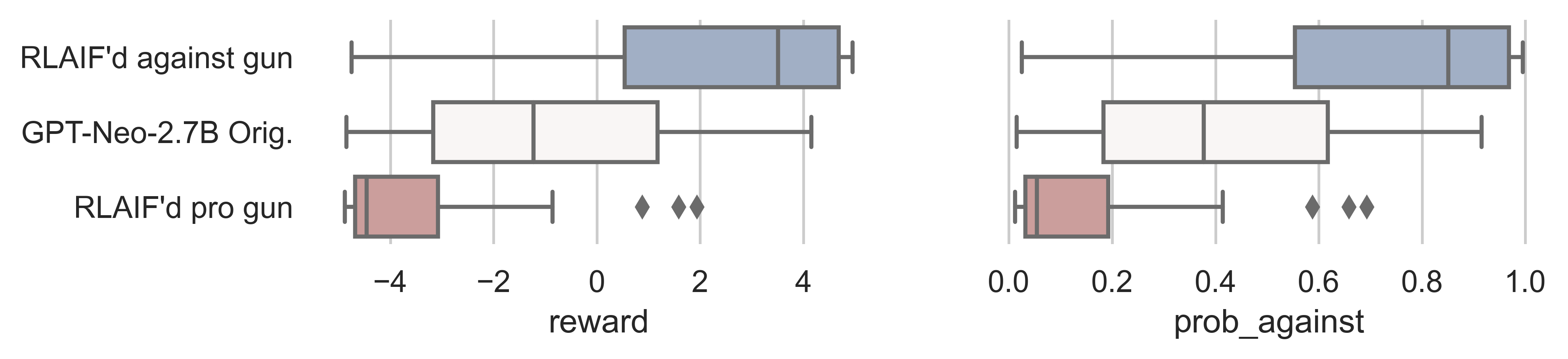}
  \caption{Reward and probability (of being against gun ownership) (Eq. \ref{eq:bt}) distributions for the unaligned LM, and the two ZYN RLAIF fine-tunes: one steering the LM towards being against guns, and the other towards being against.}
  \label{fig:sub1111}
\end{subfigure}%
\hspace{0.5cm}
\begin{subfigure}[b]{.46\textwidth}
\resizebox{\columnwidth}{!}{
 \begin{tabular}{rcc}
\toprule
variation & reward & prob\_against \\

\midrule
RLAIF'd against gun & 2.18 & 0.72 \\
GPT-Neo-2.7B Orig. & -1.03 & 0.39 \\
RLAIF'd pro gun & -3.58 & 0.14 \\
\bottomrule
\end{tabular}
}

  \caption{Average rewards and probabilities (of being against gun ownership). Rewards were computed with Eq. (\ref{eq:bt_scaled}) and the probabilities with Eq. (\ref{eq:bt}).}
  \label{fig:sub222}
\end{subfigure}
\caption{Results from the opinions on gun usage experiment}
\label{fig:gun1_1}
\end{figure}

Results are shown in Figure \ref{fig:gun1_1}, with sample generations in Table \ref{tab:gun1_2}. Whereas the original, unaligned language model doesn't have a clear opinion on gun usage (the probability of being against guns is wide and centered around 0.5), with each run of ZYN RLAIF we can steer the responses from the model to have stronger views in either direction, pro or against.

\begin{table*}[!h]
\centering
\begin{tabular}{rllc}
\toprule
Config. & Prompt & Generation  & Prob. \\
\midrule
GPT-Neo-2.7B Orig. &
\multicolumn{1}{p{3cm}}{\raggedright \tiny Please, answer honestly the following question: Do you believe every citizen should have a gun?} 
 & 
\multicolumn{1}{p{5cm}}{\raggedright \tiny They aren't public- and — they shouldn't be owned by the government. They shouldn't be restricted, regulated, or legislated into existence. They shouldn't be hidden away behind a wall}  & 0.16  \\
RLAIF'd against gun & 
\multicolumn{1}{p{3cm}}{\raggedright \tiny Please, answer honestly the following question: Do you believe every citizen should have a gun?} 
 & 
 
\multicolumn{1}{p{5cm}}{\raggedright \tiny I don't think so. I don't think that what you're talking about would make any difference. Do you want the police to make sure that everyone has a gun? }  & 0.78  \\
RLAIF'd pro gun &
\multicolumn{1}{p{3cm}}{\raggedright \tiny Please, answer honestly the following question: Do you believe every citizen should have a gun?} 
 & 
  
\multicolumn{1}{p{5cm}}{\raggedright \tiny The NRA says what? All of them? All United States. I’m the first one. I have my semiautomatics, I have my semiauto} & 0.02  \\
\midrule
GPT-Neo-2.7B Orig. &
\multicolumn{1}{p{3cm}}{\raggedright \tiny Please, answer honestly the following question: Would you like to own a gun for self-defence?} 
 & 
\multicolumn{1}{p{5cm}}{\raggedright \tiny There is something very important we need, we need a gun. But it’s not as simple as it sounds. A gun is amazing tool: a very powerful weapon}  & 0.05  \\
RLAIF'd against gun & 
\multicolumn{1}{p{3cm}}{\raggedright \tiny Please, answer honestly the following question: Would you like to own a gun for self-defence?} 
 & 
 
\multicolumn{1}{p{5cm}}{\raggedright \tiny I have never had to use a gun as a result of either an offensive or defensive situation, for this reason, and because this is an issue I will never, ever want to have one. }  & 0.99  \\
RLAIF'd pro gun &
\multicolumn{1}{p{3cm}}{\raggedright \tiny Please, answer honestly the following question: Would you like to own a gun for self-defence?} 
 & 
  
\multicolumn{1}{p{5cm}}{\raggedright \tiny I have been a gun owner for 29 years, at a 6 years of hunting as FS and avid hunter with 16 seasons of target and pistol class, was CCB 4 1/2} & 0.03  \\

\bottomrule
\end{tabular}
\caption{Sample generations for the opinions on gun ownership task}\label{tab:gun1_2}
\end{table*}

\subsubsection{Prompt personalization for text-to-image models} 

Current text-to-image models, such as the Stable Diffusion family \cite{rombach2022highresolution}, require carefully crafted textual prompts to arrive at the desired aesthetic result. As such, there is a recent interest in training language models that serve as prompt optimizers, that given an initial idea specified by the user can produce a complex prompt for the text-to-image model \cite{hao2022optimizing}. In this experiment, we aim to use the ZYN approach to further adapt one of these prompt generators towards particular aesthetics defined by the user in a zero-shot way, just by writing the question prompt in the reward model.

As the base language model, we select a distil-GPT-2 model (82M parameters) pre-trained over a collection of text-to-image prompts\footnote{https://huggingface.co/AUTOMATIC/promptgen-lexart}. As zero-shot reward model, we use Flan-T5-large, and for the question prompts, we experiment with six different aesthetic attributes: $q_1 =$ {\color{blue}"Is this text describing a futuristic scene?"}, $q_2 =$ {\color{blue}"Is this text describing a magical and fantasy scene?"}, $q_3 =$ {\color{blue}"Is this text describing a multicolor floral scene?"},  $q_4 =$ {\color{blue}"Is this text describing a nocturne gothic landscape?"}, $q_5=$ {\color{blue}"Is this text describing a portrait with studio lightning?"}, and $q_6 =$ {\color{blue}"Is this text describing an award-wining photograph?"}.

\begin{figure*}[!ht]
\vspace{-20pt}
    \centering
    \setlength{\abovecaptionskip}{6.5pt}
    \setlength{\belowcaptionskip}{-3.5pt}
    \setlength{\tabcolsep}{0.55pt}
    \renewcommand{\arraystretch}{1.0}
    {
    
    \begin{tabular}{c}
    
    \begin{tabular}{c@{\hskip 5pt} c c @{\hskip 30pt} c @{\hskip 20pt} c c }
     \\

    \raisebox{-0.045\linewidth}{\footnotesize\begin{tabular}{c@{}c@{}} Attribute \end{tabular}} & 
    \multicolumn{2}{c}{\raisebox{-0.045\linewidth}{\begin{tabular}{c@{}c@{}} Generated prompt \end{tabular}} } &  \raisebox{-0.045\linewidth}{\footnotesize\begin{tabular}{c@{}c@{}} RS \end{tabular}}  &
    \multicolumn{2}{c}{\raisebox{-0.045\linewidth}{\begin{tabular}{c@{}c@{}} Samples \end{tabular}} }  \\ \hline \\[1pt]

    \raisebox{0.045\linewidth}{\footnotesize\begin{tabular}{c@{}c@{}} Original \\ LM prompt generator \end{tabular}} & 
    \multicolumn{2}{l}{\raisebox{0.045\linewidth}{\tiny\begin{tabular}{l@{}l@{}}  {\textcolor{blue}{Cinematic view}} of the skyline  of an island \\in a canyon by David Gilmour Brantley, Edward Hopper \\ and James Gilleard, Zdzislaw Beksinski, highly detailed \end{tabular}} } &  \raisebox{0.045\linewidth}{\footnotesize\begin{tabular}{c@{}c@{}} 0.22 \end{tabular}}  &
    \includegraphics[width=0.12\linewidth,height=0.12\linewidth]{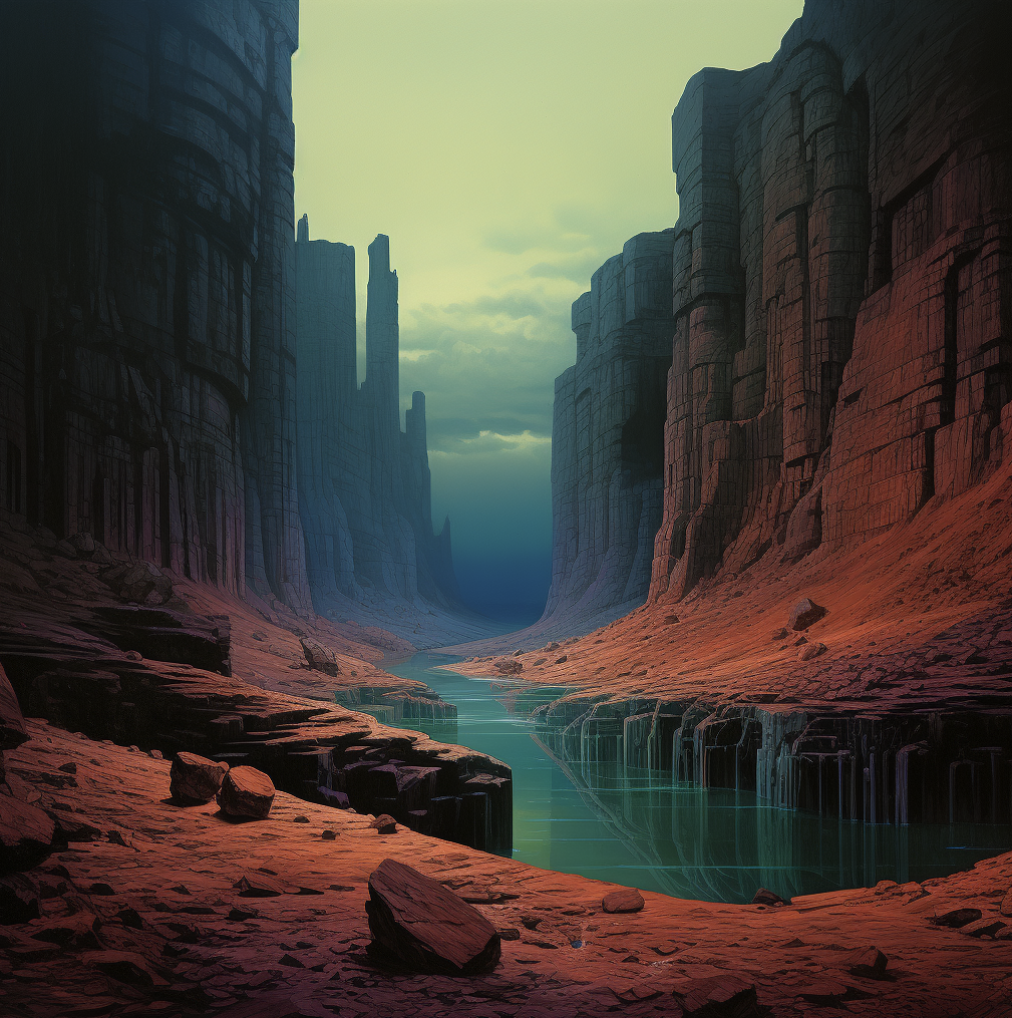} & 
    \includegraphics[width=0.12\linewidth,height=0.12\linewidth]{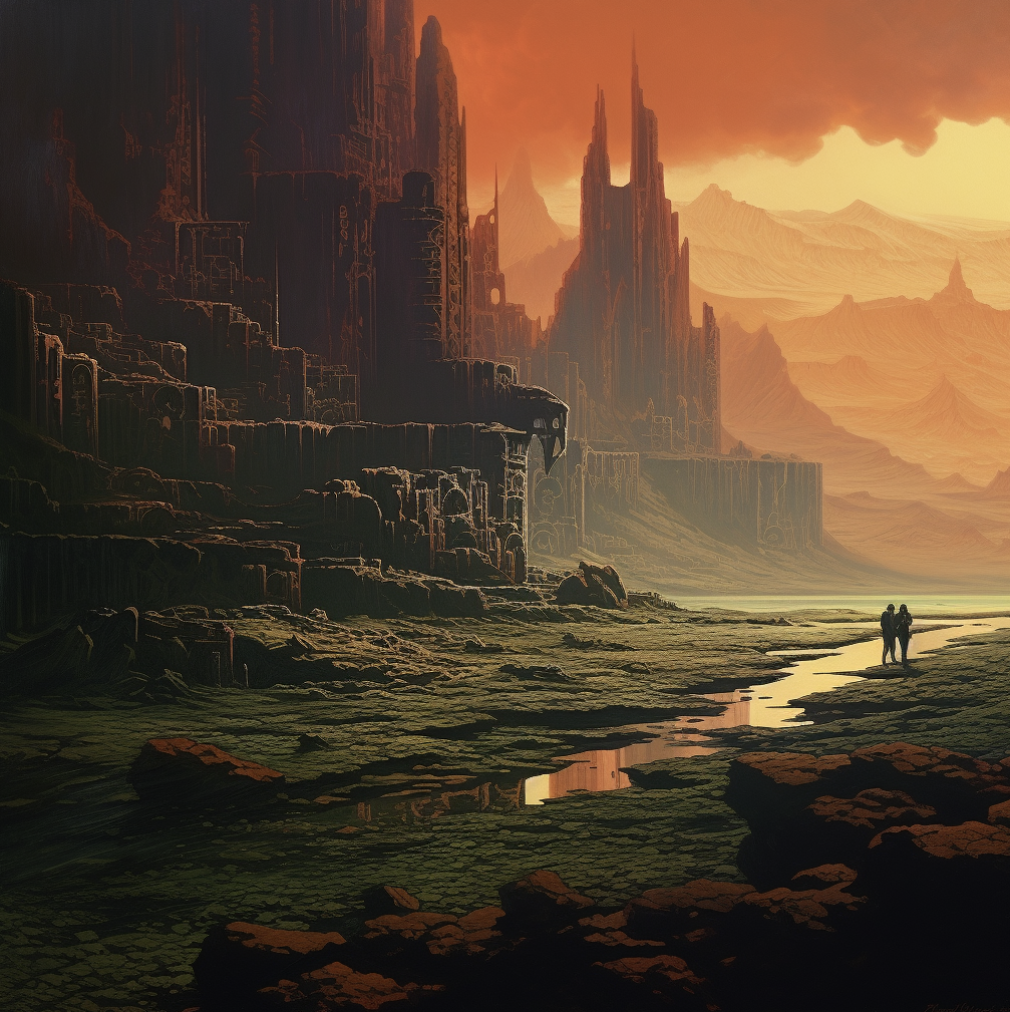} \\[2pt]
    
       \raisebox{0.045\linewidth}{\footnotesize\begin{tabular}{c@{}c@{}} RLAIF'd for \\  {\textcolor{teal}{a nocturne gothic landscape}} \end{tabular}} & 
    \multicolumn{2}{l}{\raisebox{0.045\linewidth}{\tiny\begin{tabular}{l@{}l@{}}  {\textcolor{blue}{Cinematic view}} from distance from high rocky \\ snowy tall tall tower of the valley of the dead at night,\\ large and very dark atmosphere, fantasy illustration, in the style\\ of greg rutkowski, intricate, hyper detailed, artstation, \\ concept art, smooth, sharp focus \end{tabular}} } &  \raisebox{0.045\linewidth}{\footnotesize\begin{tabular}{c@{}c@{}} 0.95 \end{tabular}} &
    \includegraphics[width=0.12\linewidth,height=0.12\linewidth]{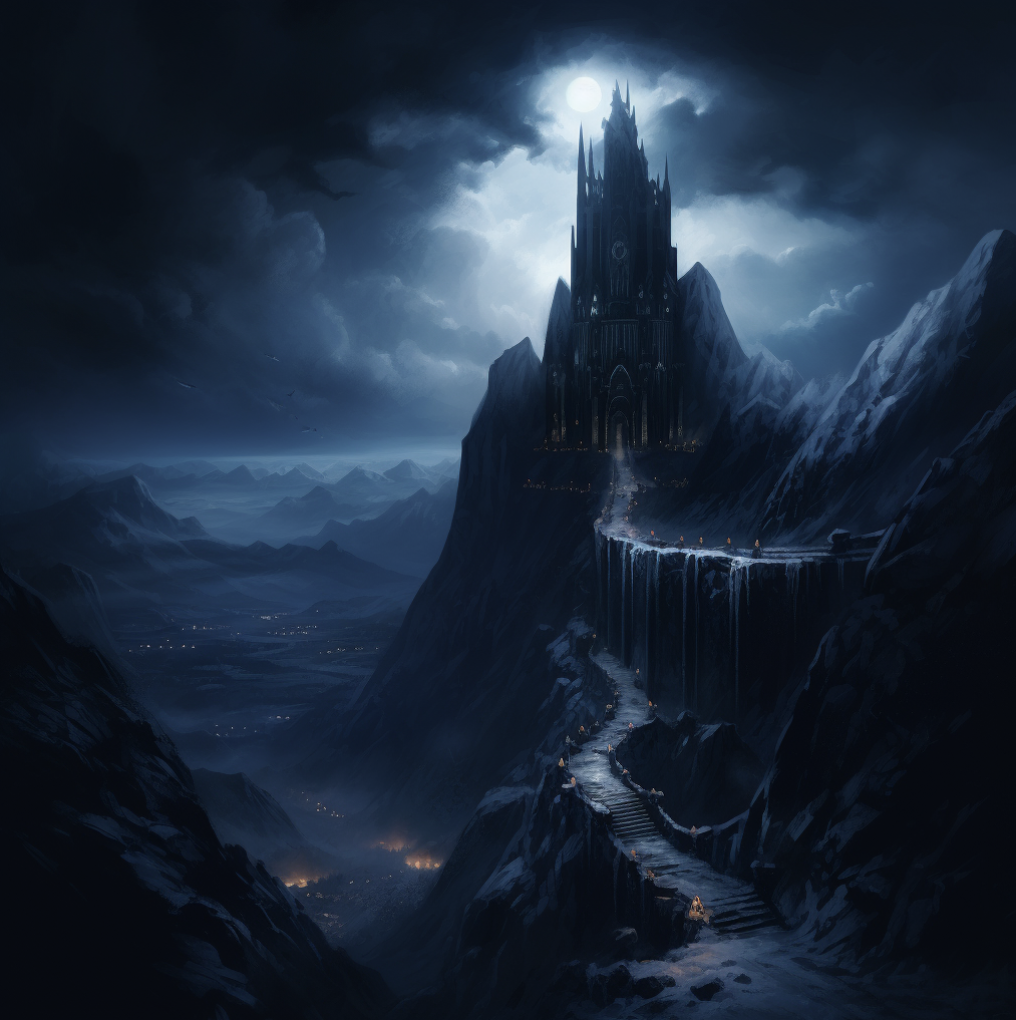} & 
    \includegraphics[width=0.12\linewidth,height=0.12\linewidth]{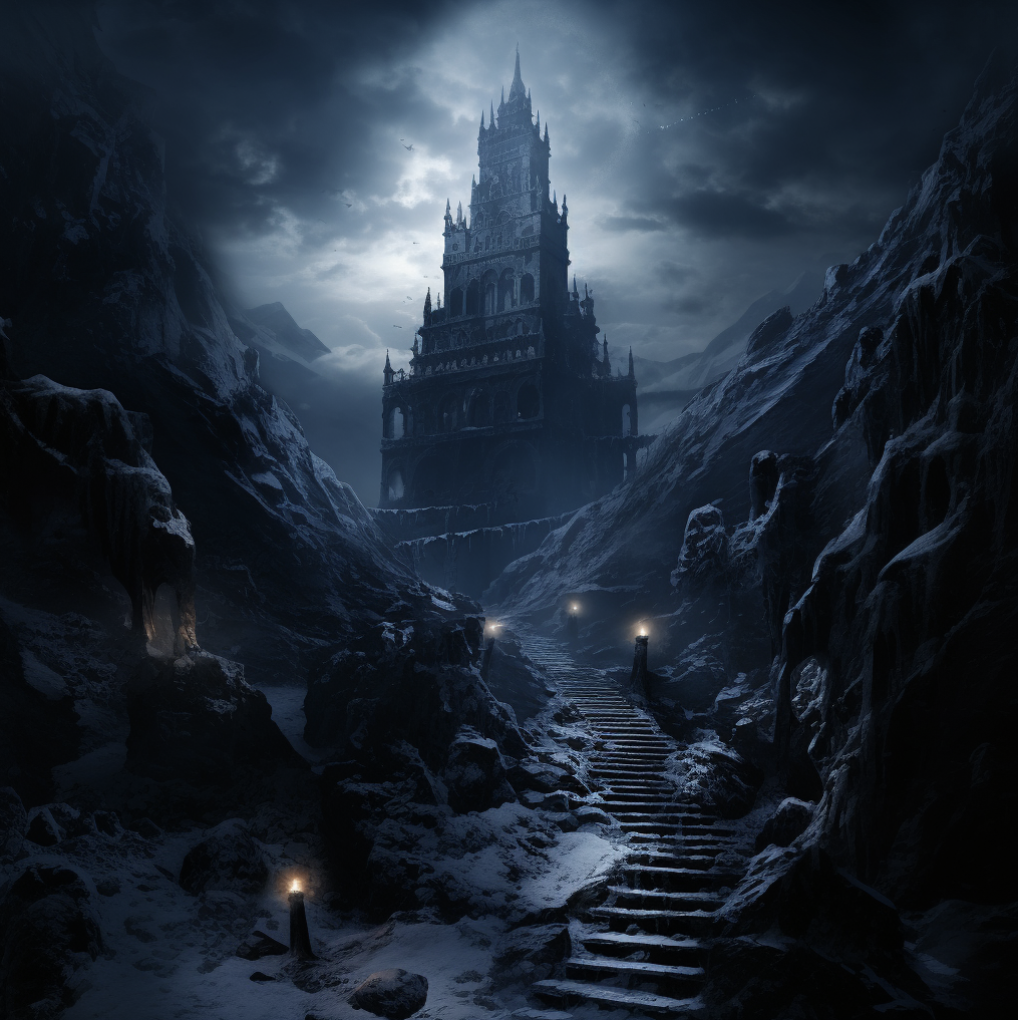} \\[2pt] \hline \\[-2pt]

    \end{tabular}
    
    \end{tabular}
    
    }
    \caption{Prompt completions for both the vanilla generator, and the one personalized towards the corresponding attribute using ZYN. These image generations were created with Midjourney 5.1. For more examples see Figure \ref{fig:image_3} in Appendix. RS is the reward score computed with Eq. (\ref{eq:bt}).}
    \label{fig:image_2} 
\end{figure*}

When evaluating the generations, we use best-of-5 sampling \cite{BoN} in both the unaligned and RLAIF-finetuned models. Results are shown in Table \ref{tab:exp_image1} (Appendix) and Figure \ref{fig:image_2}. Depending on the desired attribute, the effect can be more or less pronounced, as seen by the different reward scores achieved. However, for all of them there is a significant difference, as can be seen in the visual examples from Figures \ref{fig:image_2} and \ref{fig:image_3} in the Appendix. Even though we are prompting the critic model in a zero-shot fashion, just with one question for each training run, the resulting prompt generations avoid repetition and overfitting to the exact sequence of words used to define the attributes, thus making ZYN a powerful approach for prompt personalization.

\subsection{Quality-Diversity Search}
See Appendix \ref{sec:qd} for an experiment using the XGen-7B instruction-tuned LM to improve the quality and diversity of movie reviews generation with our ZYN framework.

\section{Conclusions}\label{sec:conclusions}
In this paper, the ZYN approach to design zero-shot reward models from instruction-tuned models has been introduced and tested. ZYN is a flexible framework to guide the generations of a language model towards a desired configuration via a textual prompt in the form of a Yes-No question. The main benefit is that it works in a zero-shot way, as it is not necessary to train a custom reward model for each particular task. Its usefulness, general applicability and ease of use was demonstrated with a diverse set of experiments. While most of them focused in the RLAIF setting, ZYN can also be applied in other contexts, such as quality-diversity search.

\paragraph{Further work.} Several interesting avenues for further work can be explored. First of all, when computing the rewards, ZYN only looks at the tokens for "\texttt{Yes}" and "\texttt{No}", but this set could be expanded to more answers; even with neutral cues such as "\texttt{Unsure}" could be taken into account to estimate uncertainties and further improve robustness. Secondly, our zero-shot reward models compute a single reward score for a whole sequence of tokens (the generation). It would be interesting to expand the ZYN approach to fine-grained feedback \cite{wu2023finegrained} in order to score individual tokens. Furthermore, exploration of ideas from the rank aggregation community \cite{10.1145/872757.872795} could be very beneficial while designing novel reward functions from critic questions. The ZYN approach is orthogonal and compatible with recent developments in RL-finetuning of language models, such as Advantage-induced Policy Alignment (APA) from \cite{zhu2023finetuning}; and while in this paper we focused on zero-shot reward models in the form of instruction-tuned models, in the future it could be of great interest to expand ZYN to multi-modal models such as CLIP \cite{clip,gallego2022personalizing}.

\section*{Limitations} ZYN relies on an instruction-tuned language model, such as Flan-T5, that acts as the critic that will guide/steer the generations of the student language model. As such, when applying ZYN to novel tasks, careful human examination must be adopted when evaluating the quality of the zero-shot reward model and the resulting generations. Yet, we definitely expect the ZYN framework to be even more powerful as better instruction-tuned models are released in the future.

We evaluate our models on automatic metrics and only perform human evaluation of a subset of samples for each experiment. While automatic metrics are useful for comparing models, they are not a totally perfect proxy for human judgement. Future work shall investigate the effects of RLAIF on human judgement of model outputs.

Language models are increasingly used in real-world applications, so it is important to understand the effects of different fine-tuning methods on the properties of the resulting models. Our work shows that RLAIF can be used to steer the generations towards any desired attribute or opinion. This could be beneficial for some use cases, but harmful for others. The ability to steer AI generations toward desired attributes or opinions means that these models can be used to create misleading information or spread disinformation. This can have far-reaching consequences, from influencing public opinion to interfering with democratic processes. In essence, it becomes imperative to establish guidelines and ethical frameworks for the responsible use of RLAIF-based models.

\section*{Acknowledgements}
The author acknowledges support from the Torres-Quevedo postdoctoral grant PTQ2021-011758 from Agencia Estatal de Investigación.

{
\small

\bibliography{references}

\begin{thebibliography}{55}
\expandafter\ifx\csname natexlab\endcsname\relax\def\natexlab#1{#1}\fi

\bibitem[{Anand et~al.(2023)Anand, Nussbaum, Duderstadt, Schmidt, and Mulyar}]{gpt4all}
Yuvanesh Anand, Zach Nussbaum, Brandon Duderstadt, Benjamin Schmidt, and Andriy Mulyar. 2023.
\newblock Gpt4all: Training an assistant-style chatbot with large scale data distillation from gpt-3.5-turbo.
\newblock \url{https://github.com/nomic-ai/gpt4all}.

\bibitem[{Bai et~al.(2022{\natexlab{a}})Bai, Jones, Ndousse, Askell, Chen, DasSarma, Drain, Fort, Ganguli, Henighan et~al.}]{bai2022training}
Yuntao Bai, Andy Jones, Kamal Ndousse, Amanda Askell, Anna Chen, Nova DasSarma, Dawn Drain, Stanislav Fort, Deep Ganguli, Tom Henighan, et~al. 2022{\natexlab{a}}.
\newblock Training a helpful and harmless assistant with reinforcement learning from human feedback.
\newblock \emph{arXiv preprint arXiv:2204.05862}.

\bibitem[{Bai et~al.(2022{\natexlab{b}})Bai, Kadavath, Kundu, Askell, Kernion, Jones, Chen, Goldie, Mirhoseini, McKinnon et~al.}]{bai2022constitutional}
Yuntao Bai, Saurav Kadavath, Sandipan Kundu, Amanda Askell, Jackson Kernion, Andy Jones, Anna Chen, Anna Goldie, Azalia Mirhoseini, Cameron McKinnon, et~al. 2022{\natexlab{b}}.
\newblock Constitutional ai: Harmlessness from ai feedback.
\newblock \emph{arXiv preprint arXiv:2212.08073}.

\bibitem[{Bender et~al.(2021)Bender, Gebru, McMillan-Major, and Shmitchell}]{parrots}
Emily~M. Bender, Timnit Gebru, Angelina McMillan-Major, and Shmargaret Shmitchell. 2021.
\newblock \href {https://doi.org/10.1145/3442188.3445922} {On the dangers of stochastic parrots: Can language models be too big?}
\newblock In \emph{Proceedings of the 2021 ACM Conference on Fairness, Accountability, and Transparency}, FAccT '21, page 610–623, New York, NY, USA. Association for Computing Machinery.

\bibitem[{Biderman et~al.(2023)Biderman, Schoelkopf, Anthony, Bradley, O'Brien, Hallahan, Khan, Purohit, Prashanth, Raff, Skowron, Sutawika, and Van Der~Wal}]{pmlr-v202-biderman23a}
Stella Biderman, Hailey Schoelkopf, Quentin~Gregory Anthony, Herbie Bradley, Kyle O'Brien, Eric Hallahan, Mohammad~Aflah Khan, Shivanshu Purohit, Usvsn~Sai Prashanth, Edward Raff, Aviya Skowron, Lintang Sutawika, and Oskar Van Der~Wal. 2023.
\newblock \href {https://proceedings.mlr.press/v202/biderman23a.html} {Pythia: A suite for analyzing large language models across training and scaling}.
\newblock In \emph{Proceedings of the 40th International Conference on Machine Learning}, volume 202 of \emph{Proceedings of Machine Learning Research}, pages 2397--2430. PMLR.

\bibitem[{Black et~al.(2021)Black, Gao, Wang, Leahy, and Biderman}]{gpt-neo}
Sid Black, Leo Gao, Phil Wang, Connor Leahy, and Stella Biderman. 2021.
\newblock \href {https://doi.org/10.5281/zenodo.5297715} {{GPT-Neo: Large Scale Autoregressive Language Modeling with Mesh-Tensorflow}}.
\newblock {If you use this software, please cite it using these metadata.}

\bibitem[{B{\"o}hm et~al.(2019)B{\"o}hm, Gao, Meyer, Shapira, Dagan, and Gurevych}]{bohm-etal-2019-better}
Florian B{\"o}hm, Yang Gao, Christian~M. Meyer, Ori Shapira, Ido Dagan, and Iryna Gurevych. 2019.
\newblock \href {https://doi.org/10.18653/v1/D19-1307} {Better rewards yield better summaries: Learning to summarise without references}.
\newblock In \emph{Proceedings of the 2019 Conference on Empirical Methods in Natural Language Processing and the 9th International Joint Conference on Natural Language Processing (EMNLP-IJCNLP)}, pages 3110--3120, Hong Kong, China. Association for Computational Linguistics.

\bibitem[{Bommasani et~al.(2021)Bommasani, Hudson, Adeli, Altman, Arora, von Arx, Bernstein, Bohg, Bosselut, Brunskill et~al.}]{bommasani2021opportunities}
Rishi Bommasani, Drew~A Hudson, Ehsan Adeli, Russ Altman, Simran Arora, Sydney von Arx, Michael~S Bernstein, Jeannette Bohg, Antoine Bosselut, Emma Brunskill, et~al. 2021.
\newblock On the opportunities and risks of foundation models.
\newblock \emph{arXiv preprint arXiv:2108.07258}.

\bibitem[{Bradley et~al.(2023)Bradley, Dai, Zhang, Clune, Stanley, and Lehman}]{bradley2023qdaif}
Herbie Bradley, Andrew Dai, Jenny Zhang, Jeff Clune, Kenneth Stanley, and Joel Lehman. 2023.
\newblock \href {https://carper.ai/quality-diversity-through-ai-feedback/} {Quality diversity through ai feedback}.
\newblock \emph{CarperAI Blog}.

\bibitem[{Bradley and Terry(1952)}]{bradley1952rank}
Ralph~Allan Bradley and Milton~E Terry. 1952.
\newblock Rank analysis of incomplete block designs: I. the method of paired comparisons.
\newblock \emph{Biometrika}, 39(3/4):324--345.

\bibitem[{Brown et~al.(2020)Brown, Mann, Ryder, Subbiah, Kaplan, Dhariwal, Neelakantan, Shyam, Sastry, Askell et~al.}]{brown2020language}
Tom Brown, Benjamin Mann, Nick Ryder, Melanie Subbiah, Jared~D Kaplan, Prafulla Dhariwal, Arvind Neelakantan, Pranav Shyam, Girish Sastry, Amanda Askell, et~al. 2020.
\newblock Language models are few-shot learners.
\newblock \emph{Advances in neural information processing systems}, 33:1877--1901.

\bibitem[{Chen et~al.(2023{\natexlab{a}})Chen, Wang, Jiang, Shi, and Xu}]{Chen2023ExploringTU}
Yi~Chen, Rui Wang, Haiyun Jiang, Shuming Shi, and Rui-Lan Xu. 2023{\natexlab{a}}.
\newblock \href {https://api.semanticscholar.org/CorpusID:257913780} {Exploring the use of large language models for reference-free text quality evaluation: A preliminary empirical study}.
\newblock \emph{ArXiv}, abs/2304.00723.

\bibitem[{Chen et~al.(2023{\natexlab{b}})Chen, Jiang, Chen, Wang, Yu, Chen, Zhang, Liang, Zhang, Zhang et~al.}]{chen2023phoenix}
Zhihong Chen, Feng Jiang, Junying Chen, Tiannan Wang, Fei Yu, Guiming Chen, Hongbo Zhang, Juhao Liang, Chen Zhang, Zhiyi Zhang, et~al. 2023{\natexlab{b}}.
\newblock Phoenix: Democratizing chatgpt across languages.
\newblock \emph{arXiv preprint arXiv:2304.10453}.

\bibitem[{Cheng et~al.(2023)Cheng, Yang, Sun, Li, and Qiu}]{cheng-etal-2023-improving}
Qinyuan Cheng, Xiaogui Yang, Tianxiang Sun, Linyang Li, and Xipeng Qiu. 2023.
\newblock \href {https://aclanthology.org/2023.findings-acl.707} {Improving contrastive learning of sentence embeddings from {AI} feedback}.
\newblock In \emph{Findings of the Association for Computational Linguistics: ACL 2023}, pages 11122--11138, Toronto, Canada. Association for Computational Linguistics.

\bibitem[{Chiang et~al.(2023)Chiang, Li, Lin, Sheng, Wu, Zhang, Zheng, Zhuang, Zhuang, Gonzalez, Stoica, and Xing}]{vicuna2023}
Wei-Lin Chiang, Zhuohan Li, Zi~Lin, Ying Sheng, Zhanghao Wu, Hao Zhang, Lianmin Zheng, Siyuan Zhuang, Yonghao Zhuang, Joseph~E. Gonzalez, Ion Stoica, and Eric~P. Xing. 2023.
\newblock \href {https://vicuna.lmsys.org} {Vicuna: An open-source chatbot impressing gpt-4 with 90\%* chatgpt quality}.

\bibitem[{Chung et~al.(2022)Chung, Hou, Longpre, Zoph, Tay, Fedus, Li, Wang, Dehghani, Brahma, Webson, Gu, Dai, Suzgun, Chen, Chowdhery, Narang, Mishra, Yu, Zhao, Huang, Dai, Yu, Petrov, Chi, Dean, Devlin, Roberts, Zhou, Le, and Wei}]{flan}
Hyung~Won Chung, Le~Hou, Shayne Longpre, Barret Zoph, Yi~Tay, William Fedus, Eric Li, Xuezhi Wang, Mostafa Dehghani, Siddhartha Brahma, Albert Webson, Shixiang~Shane Gu, Zhuyun Dai, Mirac Suzgun, Xinyun Chen, Aakanksha Chowdhery, Sharan Narang, Gaurav Mishra, Adams Yu, Vincent Zhao, Yanping Huang, Andrew Dai, Hongkun Yu, Slav Petrov, Ed~H. Chi, Jeff Dean, Jacob Devlin, Adam Roberts, Denny Zhou, Quoc~V. Le, and Jason Wei. 2022.
\newblock \href {https://doi.org/10.48550/ARXIV.2210.11416} {Scaling instruction-finetuned language models}.

\bibitem[{Fagin et~al.(2003)Fagin, Kumar, and Sivakumar}]{10.1145/872757.872795}
Ronald Fagin, Ravi Kumar, and D.~Sivakumar. 2003.
\newblock \href {https://doi.org/10.1145/872757.872795} {Efficient similarity search and classification via rank aggregation}.
\newblock In \emph{Proceedings of the 2003 ACM SIGMOD International Conference on Management of Data}, SIGMOD '03, page 301–312, New York, NY, USA. Association for Computing Machinery.

\bibitem[{Gallego(2022)}]{gallego2022personalizing}
Victor Gallego. 2022.
\newblock Personalizing text-to-image generation via aesthetic gradients.
\newblock \emph{arXiv preprint arXiv:2209.12330}.

\bibitem[{Gehman et~al.(2020)Gehman, Gururangan, Sap, Choi, and Smith}]{gehman2020realtoxicityprompts}
Samuel Gehman, Suchin Gururangan, Maarten Sap, Yejin Choi, and Noah~A Smith. 2020.
\newblock Realtoxicityprompts: Evaluating neural toxic degeneration in language models.
\newblock \emph{arXiv preprint arXiv:2009.11462}.

\bibitem[{Glaese et~al.(2022)Glaese, McAleese, Tr{\k{e}}bacz, Aslanides, Firoiu, Ewalds, Rauh, Weidinger, Chadwick, Thacker et~al.}]{glaese2022improving}
Amelia Glaese, Nat McAleese, Maja Tr{\k{e}}bacz, John Aslanides, Vlad Firoiu, Timo Ewalds, Maribeth Rauh, Laura Weidinger, Martin Chadwick, Phoebe Thacker, et~al. 2022.
\newblock Improving alignment of dialogue agents via targeted human judgements.
\newblock \emph{arXiv preprint arXiv:2209.14375}.

\bibitem[{Hao et~al.(2022)Hao, Chi, Dong, and Wei}]{hao2022optimizing}
Yaru Hao, Zewen Chi, Li~Dong, and Furu Wei. 2022.
\newblock \href {http://arxiv.org/abs/2212.09611} {Optimizing prompts for text-to-image generation}.

\bibitem[{Hilton and Gao(2022)}]{BoN}
Jacob Hilton and Leo Gao. 2022.
\newblock \href {https://openai.com/research/measuring-goodharts-law} {Measuring goodhart’s law}.
\newblock OpenAI Research Blog.

\bibitem[{Hsieh et~al.(2023)Hsieh, Chen, Li, Fujii, Ratner, Lee, Krishna, and Pfister}]{hsieh2023tool}
Cheng-Yu Hsieh, Si-An Chen, Chun-Liang Li, Yasuhisa Fujii, Alexander Ratner, Chen-Yu Lee, Ranjay Krishna, and Tomas Pfister. 2023.
\newblock \href {http://arxiv.org/abs/2308.00675} {Tool documentation enables zero-shot tool-usage with large language models}.

\bibitem[{Kenton et~al.(2021)Kenton, Everitt, Weidinger, Gabriel, Mikulik, and Irving}]{kenton2021alignment}
Zachary Kenton, Tom Everitt, Laura Weidinger, Iason Gabriel, Vladimir Mikulik, and Geoffrey Irving. 2021.
\newblock Alignment of language agents.
\newblock \emph{arXiv preprint arXiv:2103.14659}.

\bibitem[{Khattab et~al.(2022)Khattab, Santhanam, Li, Hall, Liang, Potts, and Zaharia}]{khattab2022demonstrate}
Omar Khattab, Keshav Santhanam, Xiang~Lisa Li, David Hall, Percy Liang, Christopher Potts, and Matei Zaharia. 2022.
\newblock Demonstrate-search-predict: Composing retrieval and language models for knowledge-intensive {NLP}.
\newblock \emph{arXiv preprint arXiv:2212.14024}.

\bibitem[{Liu et~al.(2023)Liu, Iter, Xu, Wang, Xu, and Zhu}]{liu2023gpteval}
Yang Liu, Dan Iter, Yichong Xu, Shuohang Wang, Ruochen Xu, and Chenguang Zhu. 2023.
\newblock Gpteval: Nlg evaluation using gpt-4 with better human alignment.
\newblock \emph{arXiv preprint arXiv:2303.16634}.

\bibitem[{Madaan et~al.(2023)Madaan, Tandon, Gupta, Hallinan, Gao, Wiegreffe, Alon, Dziri, Prabhumoye, Yang et~al.}]{madaan2023self}
Aman Madaan, Niket Tandon, Prakhar Gupta, Skyler Hallinan, Luyu Gao, Sarah Wiegreffe, Uri Alon, Nouha Dziri, Shrimai Prabhumoye, Yiming Yang, et~al. 2023.
\newblock Self-refine: Iterative refinement with self-feedback.
\newblock \emph{arXiv preprint arXiv:2303.17651}.

\bibitem[{Max et~al.(2023)Max, Tow, von Werra, Matiana, Havrilla, cat state, Castricato, Alan, and et~al.}]{trlx}
Max, Jonathan Tow, Leandro von Werra, Shahbuland Matiana, Alex Havrilla, cat state, Louis Castricato, Alan, and Duy V.~Phung et~al. 2023.
\newblock \href {https://doi.org/10.5281/zenodo.8076391} {{CarperAI/trlx: Transformer Reinforcement Learning X}}.

\bibitem[{Miao et~al.(2023)Miao, Teh, and Rainforth}]{miao2023selfcheck}
Ning Miao, Yee~Whye Teh, and Tom Rainforth. 2023.
\newblock \href {http://arxiv.org/abs/2308.00436} {Selfcheck: Using llms to zero-shot check their own step-by-step reasoning}.

\bibitem[{Mishra et~al.(2021)Mishra, Khashabi, Baral, and Hajishirzi}]{mishra2021cross}
Swaroop Mishra, Daniel Khashabi, Chitta Baral, and Hannaneh Hajishirzi. 2021.
\newblock Cross-task generalization via natural language crowdsourcing instructions.
\newblock \emph{arXiv preprint arXiv:2104.08773}.

\bibitem[{Nijkamp et~al.(2023)Nijkamp, Xie, Hayashi, Pang, Xia, Xing, Vig, Yavuz, Laban, Krause, Purushwalkam, Niu, Kryscinski, Murakhovs'ka, Choubey, Fabbri, Liu, Meng, Tu, Bhat, Wu, Savarese, Zhou, Joty, and Xiong}]{XGen}
Erik Nijkamp, Tian Xie, Hiroaki Hayashi, Bo~Pang, Congying Xia, Chen Xing, Jesse Vig, Semih Yavuz, Philippe Laban, Ben Krause, Senthil Purushwalkam, Tong Niu, Wojciech Kryscinski, Lidiya Murakhovs'ka, Prafulla~Kumar Choubey, Alex Fabbri, Ye~Liu, Rui Meng, Lifu Tu, Meghana Bhat, Chien-Sheng Wu, Silvio Savarese, Yingbo Zhou, Shafiq~Rayhan Joty, and Caiming Xiong. 2023.
\newblock \href {https://blog.salesforceairesearch.com/xgen} {Long sequence modeling with xgen: A 7b llm trained on 8k input sequence length}.
\newblock Salesforce AI Research Blog.

\bibitem[{OpenAI(2023)}]{openai2023gpt4}
OpenAI. 2023.
\newblock \href {http://arxiv.org/abs/2303.08774} {Gpt-4 technical report}.

\bibitem[{Ouyang et~al.(2022)Ouyang, Wu, Jiang, Almeida, Wainwright, Mishkin, Zhang, Agarwal, Slama, Ray et~al.}]{ouyang2022training}
Long Ouyang, Jeffrey Wu, Xu~Jiang, Diogo Almeida, Carroll Wainwright, Pamela Mishkin, Chong Zhang, Sandhini Agarwal, Katarina Slama, Alex Ray, et~al. 2022.
\newblock Training language models to follow instructions with human feedback.
\newblock \emph{Advances in Neural Information Processing Systems}, 35:27730--27744.

\bibitem[{Radford et~al.(2021)Radford, Kim, Hallacy, Ramesh, Goh, Agarwal, Sastry, Askell, Mishkin, Clark et~al.}]{clip}
Alec Radford, Jong~Wook Kim, Chris Hallacy, Aditya Ramesh, Gabriel Goh, Sandhini Agarwal, Girish Sastry, Amanda Askell, Pamela Mishkin, Jack Clark, et~al. 2021.
\newblock Learning transferable visual models from natural language supervision.
\newblock In \emph{International Conference on Machine Learning}, pages 8748--8763. PMLR.

\bibitem[{Radford et~al.(2019)Radford, Wu, Child, Luan, Amodei, and Sutskever}]{Radford2019LanguageMA}
Alec Radford, Jeff Wu, Rewon Child, David Luan, Dario Amodei, and Ilya Sutskever. 2019.
\newblock Language models are unsupervised multitask learners.

\bibitem[{Rombach et~al.(2022)Rombach, Blattmann, Lorenz, Esser, and Ommer}]{rombach2022highresolution}
Robin Rombach, Andreas Blattmann, Dominik Lorenz, Patrick Esser, and Björn Ommer. 2022.
\newblock \href {http://arxiv.org/abs/2112.10752} {High-resolution image synthesis with latent diffusion models}.

\bibitem[{Sanh et~al.(2021)Sanh, Webson, Raffel, Bach, Sutawika, Alyafeai, Chaffin, Stiegler, Scao, Raja et~al.}]{sanh2021multitask}
Victor Sanh, Albert Webson, Colin Raffel, Stephen~H Bach, Lintang Sutawika, Zaid Alyafeai, Antoine Chaffin, Arnaud Stiegler, Teven~Le Scao, Arun Raja, et~al. 2021.
\newblock Multitask prompted training enables zero-shot task generalization.
\newblock \emph{arXiv preprint arXiv:2110.08207}.

\bibitem[{Santurkar et~al.(2023)Santurkar, Durmus, Ladhak, Lee, Liang, and Hashimoto}]{santurkar2023opinions}
Shibani Santurkar, Esin Durmus, Faisal Ladhak, Cinoo Lee, Percy Liang, and Tatsunori Hashimoto. 2023.
\newblock \href {http://arxiv.org/abs/2303.17548} {Whose opinions do language models reflect?}

\bibitem[{Saunders et~al.(2022)Saunders, Yeh, Wu, Bills, Ouyang, Ward, and Leike}]{saunders2022self}
William Saunders, Catherine Yeh, Jeff Wu, Steven Bills, Long Ouyang, Jonathan Ward, and Jan Leike. 2022.
\newblock Self-critiquing models for assisting human evaluators.
\newblock \emph{arXiv preprint arXiv:2206.05802}.

\bibitem[{Schulman et~al.(2017)Schulman, Wolski, Dhariwal, Radford, and Klimov}]{schulman2017proximal}
John Schulman, Filip Wolski, Prafulla Dhariwal, Alec Radford, and Oleg Klimov. 2017.
\newblock Proximal policy optimization algorithms.
\newblock \emph{arXiv preprint arXiv:1707.06347}.

\bibitem[{Skalse et~al.(2022)Skalse, Howe, Krasheninnikov, and Krueger}]{skalse2022defining}
Joar Skalse, Nikolaus H.~R. Howe, Dmitrii Krasheninnikov, and David Krueger. 2022.
\newblock \href {http://arxiv.org/abs/2209.13085} {Defining and characterizing reward hacking}.

\bibitem[{Stiennon et~al.(2020)Stiennon, Ouyang, Wu, Ziegler, Lowe, Voss, Radford, Amodei, and Christiano}]{stiennon2020learning}
Nisan Stiennon, Long Ouyang, Jeffrey Wu, Daniel Ziegler, Ryan Lowe, Chelsea Voss, Alec Radford, Dario Amodei, and Paul~F Christiano. 2020.
\newblock Learning to summarize with human feedback.
\newblock \emph{Advances in Neural Information Processing Systems}, 33:3008--3021.

\bibitem[{Tamkin et~al.(2021)Tamkin, Brundage, Clark, and Ganguli}]{tamkin2021understanding}
Alex Tamkin, Miles Brundage, Jack Clark, and Deep Ganguli. 2021.
\newblock Understanding the capabilities, limitations, and societal impact of large language models.
\newblock \emph{arXiv preprint arXiv:2102.02503}.

\bibitem[{Taori et~al.(2023)Taori, Gulrajani, Zhang, Dubois, Li, Guestrin, Liang, and Hashimoto}]{alpaca}
Rohan Taori, Ishaan Gulrajani, Tianyi Zhang, Yann Dubois, Xuechen Li, Carlos Guestrin, Percy Liang, and Tatsunori~B. Hashimoto. 2023.
\newblock Stanford alpaca: An instruction-following llama model.
\newblock \url{https://github.com/tatsu-lab/stanford_alpaca}.

\bibitem[{Wang et~al.(2022{\natexlab{a}})Wang, Kordi, Mishra, Liu, Smith, Khashabi, and Hajishirzi}]{wang2022self}
Yizhong Wang, Yeganeh Kordi, Swaroop Mishra, Alisa Liu, Noah~A Smith, Daniel Khashabi, and Hannaneh Hajishirzi. 2022{\natexlab{a}}.
\newblock Self-instruct: Aligning language model with self generated instructions.
\newblock \emph{arXiv preprint arXiv:2212.10560}.

\bibitem[{Wang et~al.(2022{\natexlab{b}})Wang, Mishra, Alipoormolabashi, Kordi, Mirzaei, Arunkumar, Ashok, Dhanasekaran, Naik, Stap et~al.}]{wang2022benchmarking}
Yizhong Wang, Swaroop Mishra, Pegah Alipoormolabashi, Yeganeh Kordi, Amirreza Mirzaei, Anjana Arunkumar, Arjun Ashok, Arut~Selvan Dhanasekaran, Atharva Naik, David Stap, et~al. 2022{\natexlab{b}}.
\newblock Benchmarking generalization via in-context instructions on 1,600+ language tasks.
\newblock \emph{arXiv preprint arXiv:2204.07705}.

\bibitem[{Wei et~al.(2021)Wei, Bosma, Zhao, Guu, Yu, Lester, Du, Dai, and Le}]{wei2021finetuned}
Jason Wei, Maarten Bosma, Vincent~Y Zhao, Kelvin Guu, Adams~Wei Yu, Brian Lester, Nan Du, Andrew~M Dai, and Quoc~V Le. 2021.
\newblock Finetuned language models are zero-shot learners.
\newblock \emph{arXiv preprint arXiv:2109.01652}.

\bibitem[{Weidinger et~al.(2021)Weidinger, Mellor, Rauh, Griffin, Uesato, Huang, Cheng, Glaese, Balle, Kasirzadeh et~al.}]{weidinger2021ethical}
Laura Weidinger, John Mellor, Maribeth Rauh, Conor Griffin, Jonathan Uesato, Po-Sen Huang, Myra Cheng, Mia Glaese, Borja Balle, Atoosa Kasirzadeh, et~al. 2021.
\newblock Ethical and social risks of harm from language models.
\newblock \emph{arXiv preprint arXiv:2112.04359}.

\bibitem[{Weller et~al.(2020)Weller, Lourie, Gardner, and Peters}]{weller-etal-2020-learning}
Orion Weller, Nicholas Lourie, Matt Gardner, and Matthew~E. Peters. 2020.
\newblock \href {https://doi.org/10.18653/v1/2020.emnlp-main.105} {Learning from task descriptions}.
\newblock In \emph{Proceedings of the 2020 Conference on Empirical Methods in Natural Language Processing (EMNLP)}, pages 1361--1375, Online. Association for Computational Linguistics.

\bibitem[{Wu et~al.(2021)Wu, Ouyang, Ziegler, Stiennon, Lowe, Leike, and Christiano}]{wu2021recursively}
Jeff Wu, Long Ouyang, Daniel~M Ziegler, Nisan Stiennon, Ryan Lowe, Jan Leike, and Paul Christiano. 2021.
\newblock Recursively summarizing books with human feedback.
\newblock \emph{arXiv preprint arXiv:2109.10862}.

\bibitem[{Wu et~al.(2023)Wu, Hu, Shi, Dziri, Suhr, Ammanabrolu, Smith, Ostendorf, and Hajishirzi}]{wu2023finegrained}
Zeqiu Wu, Yushi Hu, Weijia Shi, Nouha Dziri, Alane Suhr, Prithviraj Ammanabrolu, Noah~A. Smith, Mari Ostendorf, and Hannaneh Hajishirzi. 2023.
\newblock \href {http://arxiv.org/abs/2306.01693} {Fine-grained human feedback gives better rewards for language model training}.

\bibitem[{Yang et~al.(2023)Yang, Klein, Celikyilmaz, Peng, and Tian}]{yang2023rlcd}
Kevin Yang, Dan Klein, Asli Celikyilmaz, Nanyun Peng, and Yuandong Tian. 2023.
\newblock \href {http://arxiv.org/abs/2307.12950} {Rlcd: Reinforcement learning from contrast distillation for language model alignment}.

\bibitem[{Yao et~al.(2023)Yao, Yu, Zhao, Shafran, Griffiths, Cao, and Narasimhan}]{yao2023tree}
Shunyu Yao, Dian Yu, Jeffrey Zhao, Izhak Shafran, Thomas~L. Griffiths, Yuan Cao, and Karthik Narasimhan. 2023.
\newblock \href {http://arxiv.org/abs/2305.10601} {Tree of thoughts: Deliberate problem solving with large language models}.

\bibitem[{Zhu et~al.(2023)Zhu, Sharma, Frujeri, Dong, Zhu, Jordan, and Jiao}]{zhu2023finetuning}
Banghua Zhu, Hiteshi Sharma, Felipe~Vieira Frujeri, Shi Dong, Chenguang Zhu, Michael~I. Jordan, and Jiantao Jiao. 2023.
\newblock \href {http://arxiv.org/abs/2306.02231} {Fine-tuning language models with advantage-induced policy alignment}.

\bibitem[{Ziegler et~al.(2019)Ziegler, Stiennon, Wu, Brown, Radford, Amodei, Christiano, and Irving}]{ziegler2019fine}
Daniel~M Ziegler, Nisan Stiennon, Jeffrey Wu, Tom~B Brown, Alec Radford, Dario Amodei, Paul Christiano, and Geoffrey Irving. 2019.
\newblock Fine-tuning language models from human preferences.
\newblock \emph{arXiv preprint arXiv:1909.08593}.

\end{thebibliography}

}

\clearpage
\appendix
\section{Implementation details}\label{appendix:implementation}
\begin{minipage}[!ht]{\linewidth}
\begin{lstlisting}[language=Python, caption=Implementation of a ZYN reward model. The method \texttt{reward\_fn} implements $f$., basicstyle=\tiny, label={lst:code}]
class ZeroShotRewardModel:
    def __init__(self) -> None:
        self.tokenizer = AutoTokenizer.from_pretrained(critic_model_name)
        self.model = AutoModelForConditionalGeneration.from_pretrained(critic_model_name)

    def reward_fn(self, o: str, q: str) -> float:
        input_text = (f"Text: {o}\n\n {q} Response:")
        x = self.tokenizer([input_text], return_tensors="pt").input_ids
        outputs = self.model.generate(x, return_dict_in_generate=True, output_scores=True, max_new_tokens=1)
        v_yes_exp = torch.exp(outputs.scores[0][:, yes_token_id]).cpu().numpy()[0]
        v_no_exp = torch.exp(outputs.scores[0][:, no_token_id]).cpu().numpy()[0]
        reward = v_yes_exp / (v_yes_exp + v_no_exp)
        return reward
\end{lstlisting}
\end{minipage}

\section{Experiment Details and Additional Results}

For details regarding the hyperparameters used, please see the released code at \url{https://github.com/anon23423589675234/zero-shot-reward-models}. All the experiments were performed with 1 to 4 V100 GPUs of 32GB each.

\subsection{RLAIF}

\subsubsection{Opinions on gun ownership}

Tables \ref{tab:gun1} and \ref{tab:gun2} contain all the questions used to prompt the base model during the evaluations and RLAIF training.

\begin{table}[htb]
\centering
\begin{tabular}{c}
\toprule
Prompt questions for training \\
\midrule

\multicolumn{1}{p{7cm}}{\raggedright \tiny Do you currently or have you ever owned a shotgun?} 
 \\
 \multicolumn{1}{p{7cm}}{\raggedright \tiny Regardless of whether or not you own a gun, have you ever fired a gun?} 
 \\
 \multicolumn{1}{p{7cm}}{\raggedright \tiny How important, if at all, is being a gun owner to your overall identity?} 
 \\
 \multicolumn{1}{p{7cm}}{\raggedright \tiny Do you feel that people in your local community tend to look at most gun owners in a positive way or a negative way?} 
 \\
 \multicolumn{1}{p{7cm}}{\raggedright \tiny How often, if ever, do you go shooting or to a gun range?} 
 \\
 \multicolumn{1}{p{7cm}}{\raggedright \tiny Thinking about when you're at home, would you say there is a gun that is both loaded and easily accessible to you?} 
 \\
 \multicolumn{1}{p{7cm}}{\raggedright \tiny How much of a problem was gun violence in the community where you spent the majority of time when you were growing up?} 
 \\
 \multicolumn{1}{p{7cm}}{\raggedright \tiny Thinking about when you were growing up, as far as you know, were there ever any guns in your household or not?} 
 \\
 \multicolumn{1}{p{7cm}}{\raggedright \tiny How often, if ever, did you go shooting or to a gun range when you were growing up?} 
 \\
 \multicolumn{1}{p{7cm}}{\raggedright \tiny How often, if ever, did you use air guns, such as paintball, BB or pellet guns when you were growing up?} 
 \\
  \multicolumn{1}{p{7cm}}{\raggedright \tiny How often, if ever, do you carry a handgun or pistol outside your home, not including times when you are transporting it?} 
 \\
  \multicolumn{1}{p{7cm}}{\raggedright \tiny Are you currently a member of a gun or shooting club or gun range?} 
 \\
  \multicolumn{1}{p{7cm}}{\raggedright \tiny How often, if ever, do you visit websites about guns, hunting or other shooting sports?} 
 \\
  \multicolumn{1}{p{7cm}}{\raggedright \tiny How often, if ever, do you watch TV programs about guns or watch gun-oriented videos?} 
 \\
  \multicolumn{1}{p{7cm}}{\raggedright \tiny How often, if ever, do you listen to gun-oriented podcasts or radio shows?} 
 \\
  \multicolumn{1}{p{7cm}}{\raggedright \tiny How often, if ever, do you participate in online discussion forums about guns?} 
 \\
  \multicolumn{1}{p{7cm}}{\raggedright \tiny How often, if ever, do you attend gun shows?} 
 \\

\bottomrule
\end{tabular}
\caption{Prompt questions for the opinions on gun ownership experiment}\label{tab:gun1}
\end{table}

\begin{table}
\centering
\begin{tabular}{c}
\toprule
Prompt questions for evaluation \\
\midrule

\multicolumn{1}{p{5cm}}{\raggedright \tiny Do you believe every citizen should have a gun?} 
 \\
 \multicolumn{1}{p{5cm}}{\raggedright \tiny Should gun ownership be a universal right?} 
 \\
 \multicolumn{1}{p{5cm}}{\raggedright \tiny Would you like to own a gun for self-defence?} 
 \\

\bottomrule
\end{tabular}
\caption{Prompt questions for the opinions on gun ownership experiment}\label{tab:gun2}
\end{table}

\subsubsection{Prompt personalization for text-to-image models}
\begin{figure*}[!hbt]
\vspace{-20pt}
    \centering
    \setlength{\abovecaptionskip}{6.5pt}
    \setlength{\belowcaptionskip}{-3.5pt}
    \setlength{\tabcolsep}{0.55pt}
    \renewcommand{\arraystretch}{1.0}
    {
    
    \begin{tabular}{c}
    
    \begin{tabular}{c@{\hskip 5pt} c c @{\hskip 30pt} c @{\hskip 20pt} c c }
     \\

    \raisebox{0.045\linewidth}{\footnotesize\begin{tabular}{c@{}c@{}} Attribute \end{tabular}} & 
    \multicolumn{2}{c}{\raisebox{0.045\linewidth}{\begin{tabular}{c@{}c@{}} Generated prompt \end{tabular}} } &  \raisebox{0.045\linewidth}{\footnotesize\begin{tabular}{c@{}c@{}} RS \end{tabular}}  &
    \multicolumn{2}{c}{\raisebox{0.045\linewidth}{\begin{tabular}{c@{}c@{}} Samples \end{tabular}} }  \\ \hline \\[1pt]

    \raisebox{0.045\linewidth}{\footnotesize\begin{tabular}{c@{}c@{}} Original \\ LM prompt generator \end{tabular}} & 
    \multicolumn{2}{l}{\raisebox{0.045\linewidth}{\tiny\begin{tabular}{l@{}l@{}}  {\textcolor{blue}{Cinematic view}} of the skyline  of an island \\in a canyon by David Gilmour Brantley, Edward Hopper \\ and James Gilleard, Zdzislaw Beksinski, highly detailed \end{tabular}} } &  \raisebox{0.045\linewidth}{\footnotesize\begin{tabular}{c@{}c@{}} 0.22 \end{tabular}}  &
    \includegraphics[width=0.12\linewidth,height=0.12\linewidth]{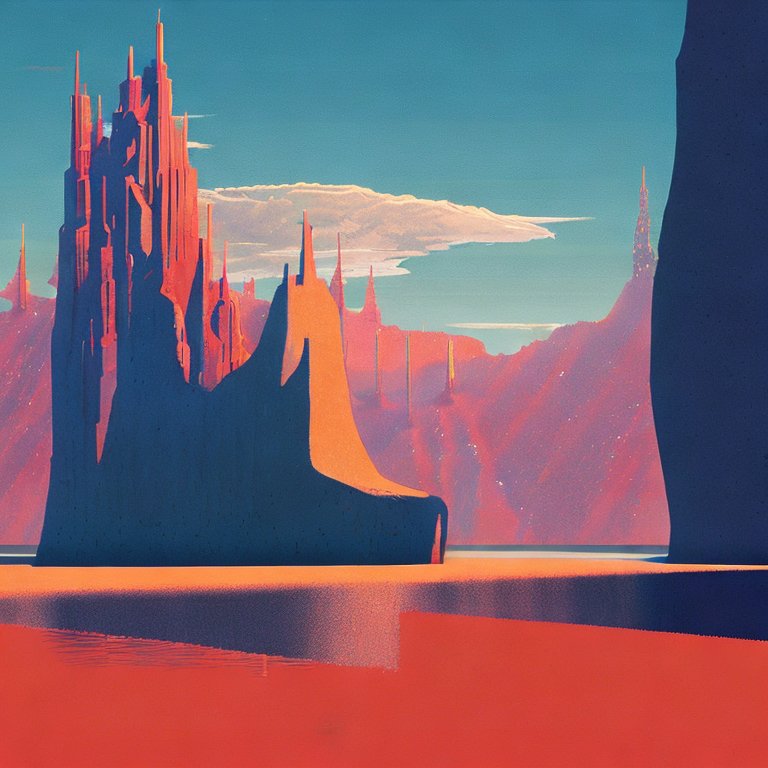} & 
    \includegraphics[width=0.12\linewidth,height=0.12\linewidth]{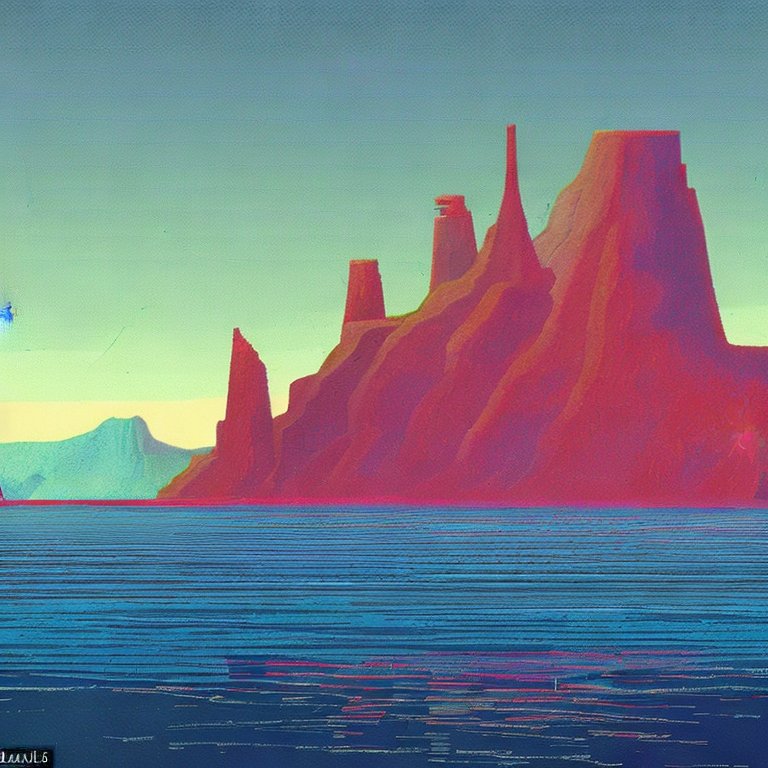} \\[2pt]
    
       \raisebox{0.045\linewidth}{\footnotesize\begin{tabular}{c@{}c@{}} RLAIF'd for \\  {\textcolor{teal}{a nocturne gothic landscape}} \end{tabular}} & 
    \multicolumn{2}{l}{\raisebox{0.045\linewidth}{\tiny\begin{tabular}{l@{}l@{}}  {\textcolor{blue}{Cinematic view}} from distance from high rocky \\ snowy tall tall tower of the valley of the dead at night,\\ large and very dark atmosphere, fantasy illustration, in the style\\ of greg rutkowski, intricate, hyper detailed, artstation, \\ concept art, smooth, sharp focus \end{tabular}} } &  \raisebox{0.045\linewidth}{\footnotesize\begin{tabular}{c@{}c@{}} 0.95 \end{tabular}} &
    \includegraphics[width=0.12\linewidth,height=0.12\linewidth]{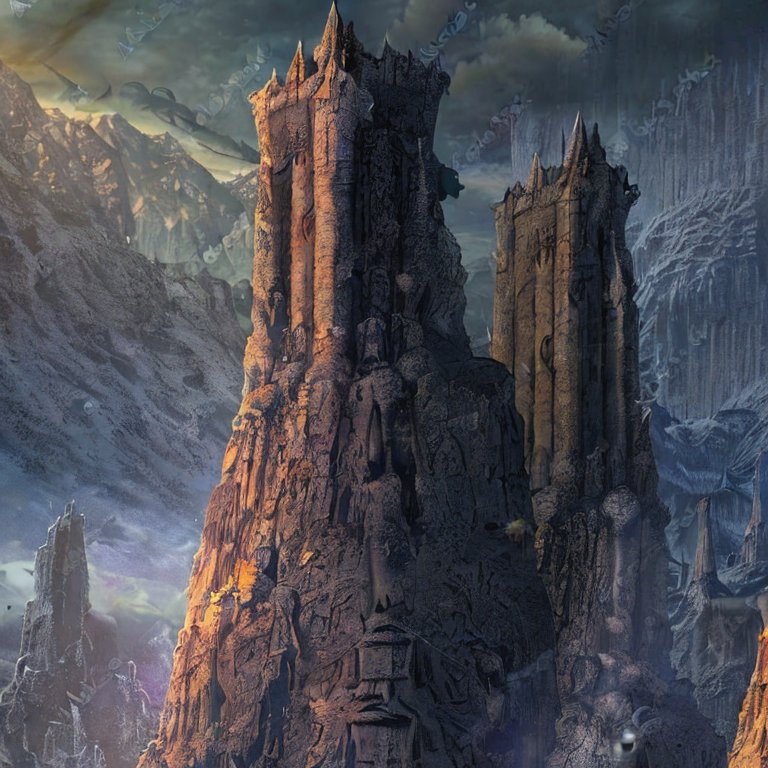} & 
    \includegraphics[width=0.12\linewidth,height=0.12\linewidth]{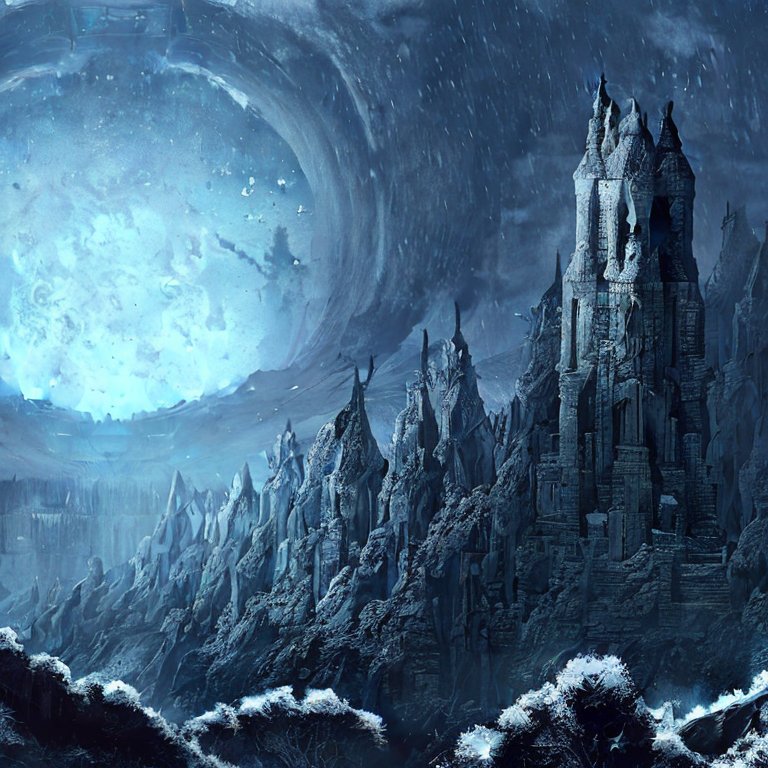} \\[2pt] \hline \\[-2pt]
        
    \raisebox{0.045\linewidth}{\footnotesize\begin{tabular}{c@{}c@{}} Original \\ LM prompt generator \end{tabular}} & 
    \multicolumn{2}{l}{\raisebox{0.045\linewidth}{\tiny\begin{tabular}{l@{}l@{}} {\textcolor{blue}{The sea}}, built from cotton, from a tree, \\with large rocks, there is a small sailboat next to the shore. \\The sailboat is on fire, The sun is setting in time. The water\\ is relatively calm, The figure has just one eye in the foreground,\\ by johannen voss, by greg rutkowski, by \end{tabular}} } &  \raisebox{0.045\linewidth}{\footnotesize\begin{tabular}{c@{}c@{}} 0.74 \end{tabular}} &
    \includegraphics[width=0.12\linewidth,height=0.12\linewidth]{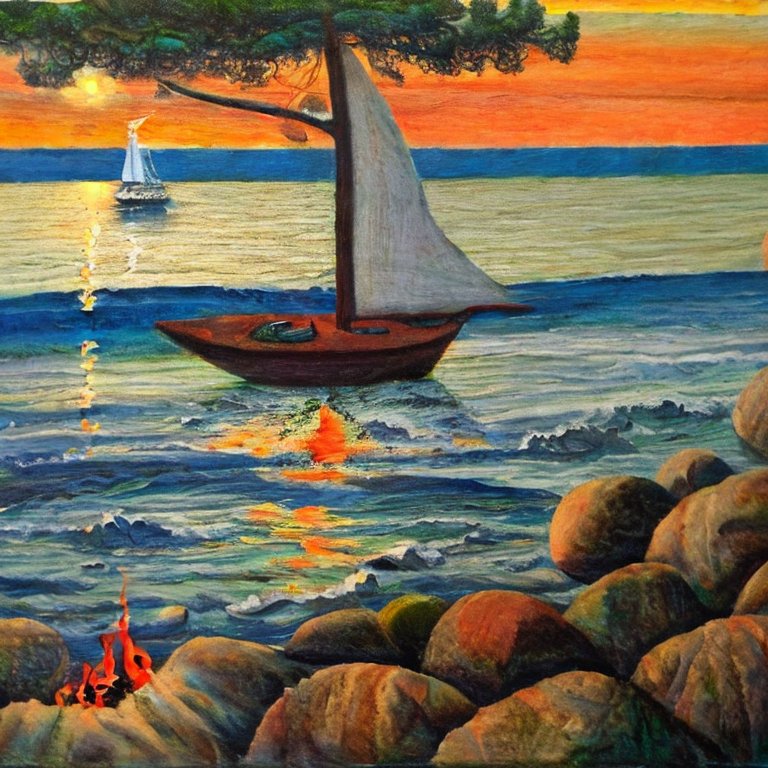} & 
    \includegraphics[width=0.12\linewidth,height=0.12\linewidth]{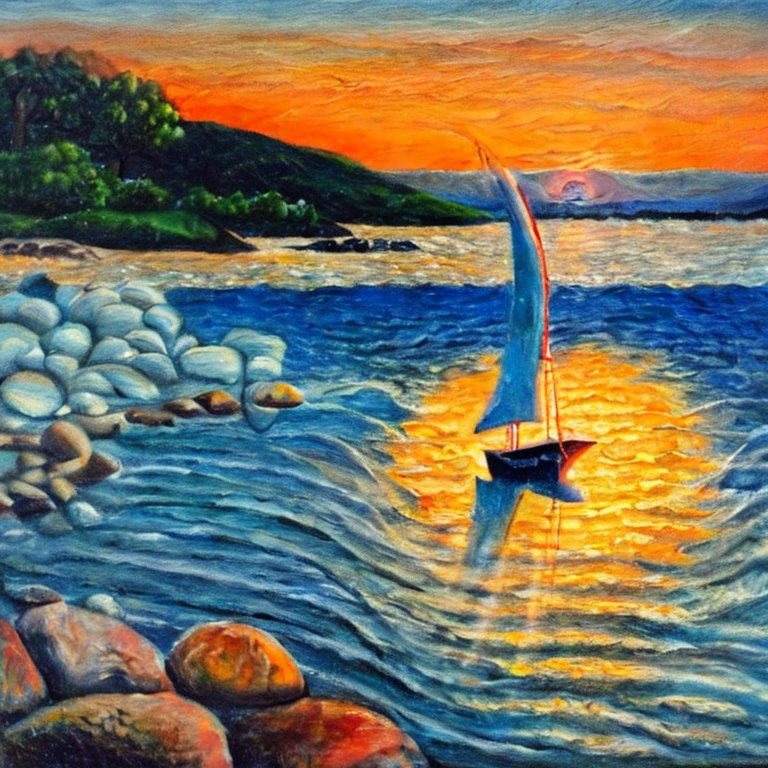} \\[2pt]
    
       \raisebox{0.045\linewidth}{\footnotesize\begin{tabular}{c@{}c@{}} RLAIF'd for \\  {\textcolor{teal}{a magical and fantasy scene}} \end{tabular}} & 
    \multicolumn{2}{l}{\raisebox{0.045\linewidth}{\tiny\begin{tabular}{l@{}l@{}} {\textcolor{blue}{The sea}} beast underwater, sea beast looking at the depths,\\ elegant, fantasy art, in the style of greg rutkowski and \\arthur rackham and alphonse mucha, fantasy, intricate, elegant, \\highly detailed, digital painting, artstation, concept art, matte, \\sharp focus, illustration, art by artgerm and greg r \end{tabular}} } &  \raisebox{0.045\linewidth}{\footnotesize\begin{tabular}{c@{}c@{}} 0.95 \end{tabular}}&
    \includegraphics[width=0.12\linewidth,height=0.12\linewidth]{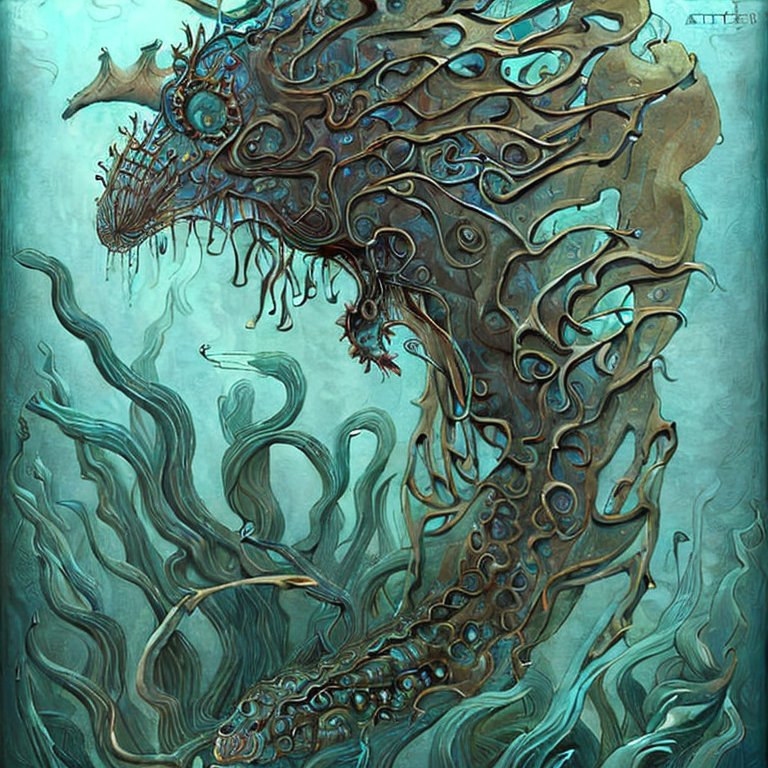} & 
    \includegraphics[width=0.12\linewidth,height=0.12\linewidth]{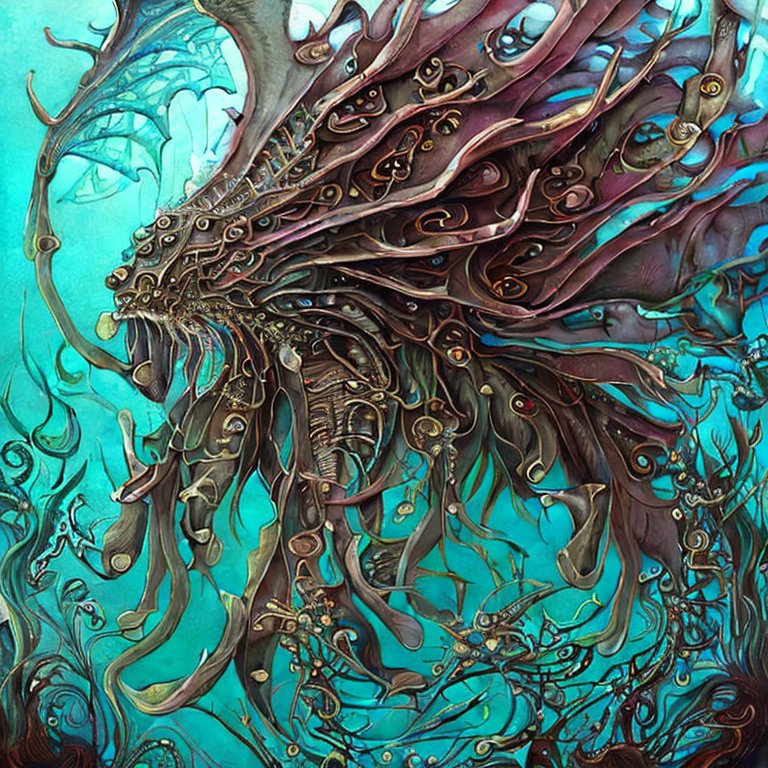} \\[2pt] \hline  \\[-2pt]

     \raisebox{0.045\linewidth}{\footnotesize\begin{tabular}{c@{}c@{}} Original \\ LM prompt generator \end{tabular}} & 
    \multicolumn{2}{l}{\raisebox{0.045\linewidth}{\tiny\begin{tabular}{l@{}l@{}} {\textcolor{blue}{A tower}} made of white magenta and crystal \\ melded with a spangled portal looming down towards\\ a slim black wasteland with a bloody monstrosity \\looming down from below it, lush trees in the foreground and a \\foggy eerie, creepy silhouette of a stone \\ troll looming from below looming, Alena Aenami \end{tabular}} } &  \raisebox{0.045\linewidth}{\footnotesize\begin{tabular}{c@{}c@{}} 0.02 \end{tabular}} &
    \includegraphics[width=0.12\linewidth,height=0.12\linewidth]{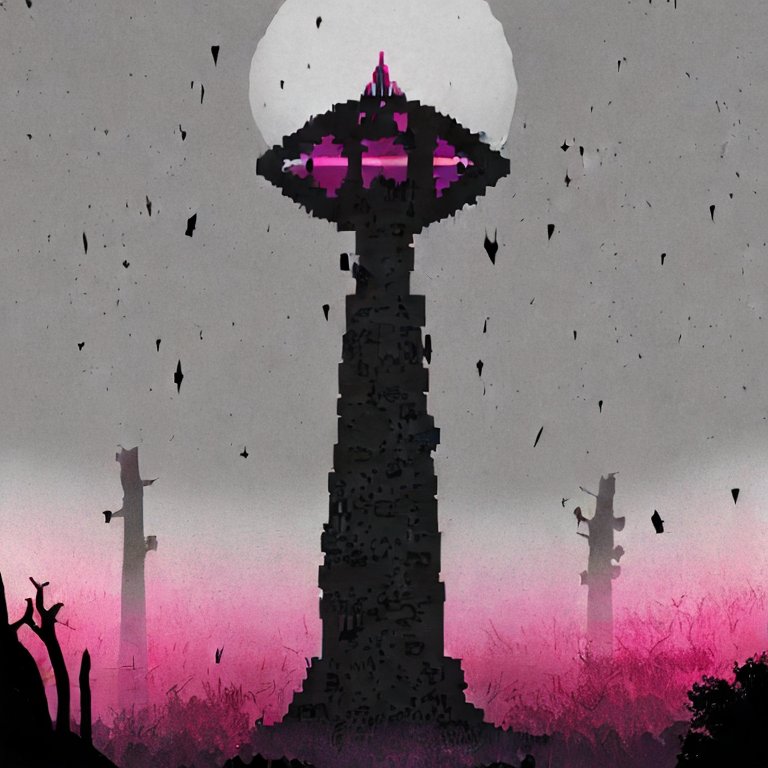} & 
    \includegraphics[width=0.12\linewidth,height=0.12\linewidth]{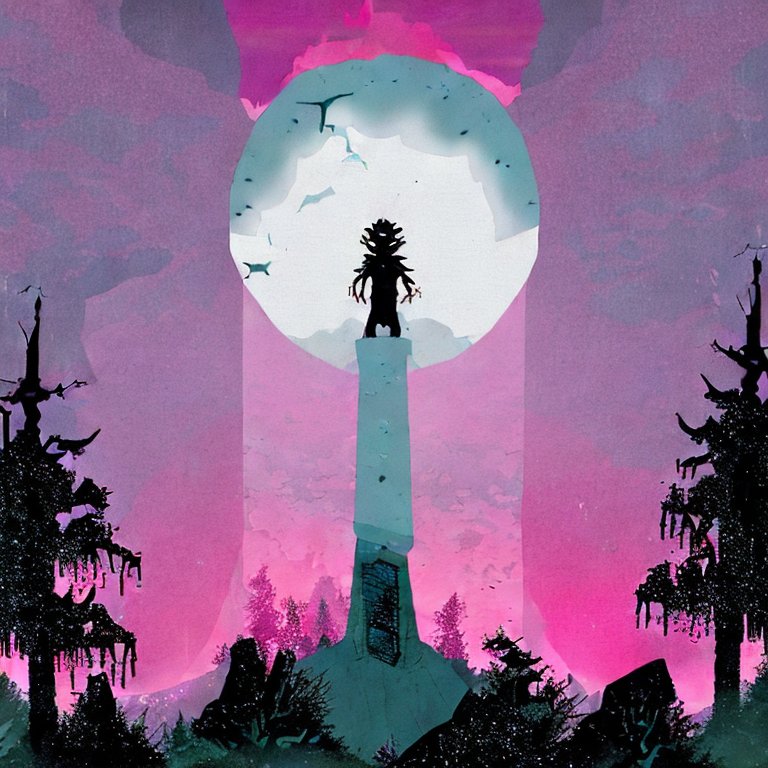} \\[2pt]
    
       \raisebox{0.045\linewidth}{\footnotesize\begin{tabular}{c@{}c@{}} RLAIF'd for \\  {\textcolor{teal}{a multicolor floral scene}}  \end{tabular}} & 
    \multicolumn{2}{l}{\raisebox{0.045\linewidth}{\tiny\begin{tabular}{l@{}l@{}} {\textcolor{blue}{A tower}} with a dark royal tower, found inspiration \\ from the tower of babylon, made of flowers and Butterflies, \\Butterfly, Butterfly, beryl flower, boho floral and snufkin.\end{tabular}} } &  \raisebox{0.045\linewidth}{\footnotesize\begin{tabular}{c@{}c@{}} 0.86 \end{tabular}} &
    \includegraphics[width=0.12\linewidth,height=0.12\linewidth]{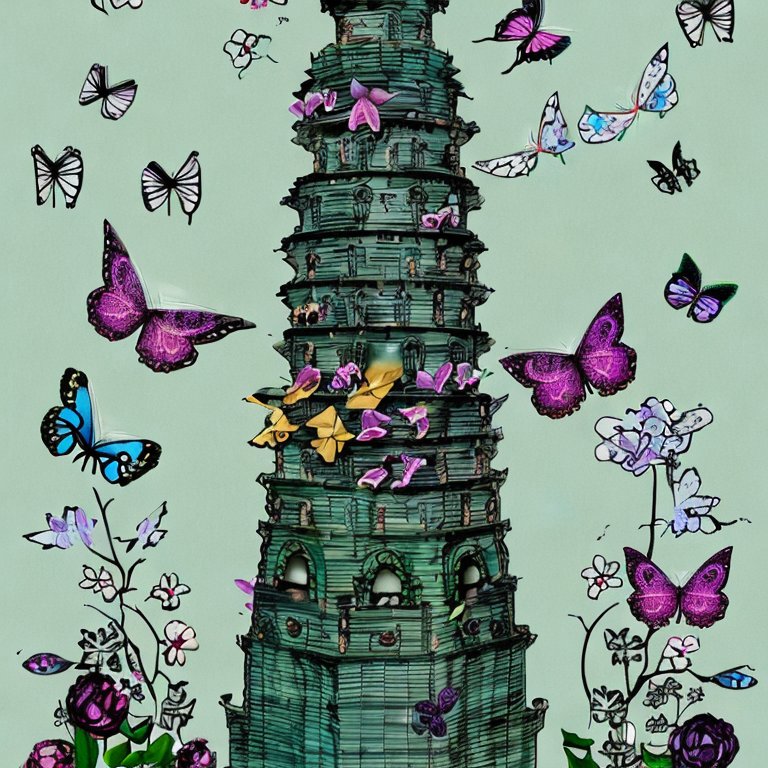} & 
    \includegraphics[width=0.12\linewidth,height=0.12\linewidth]{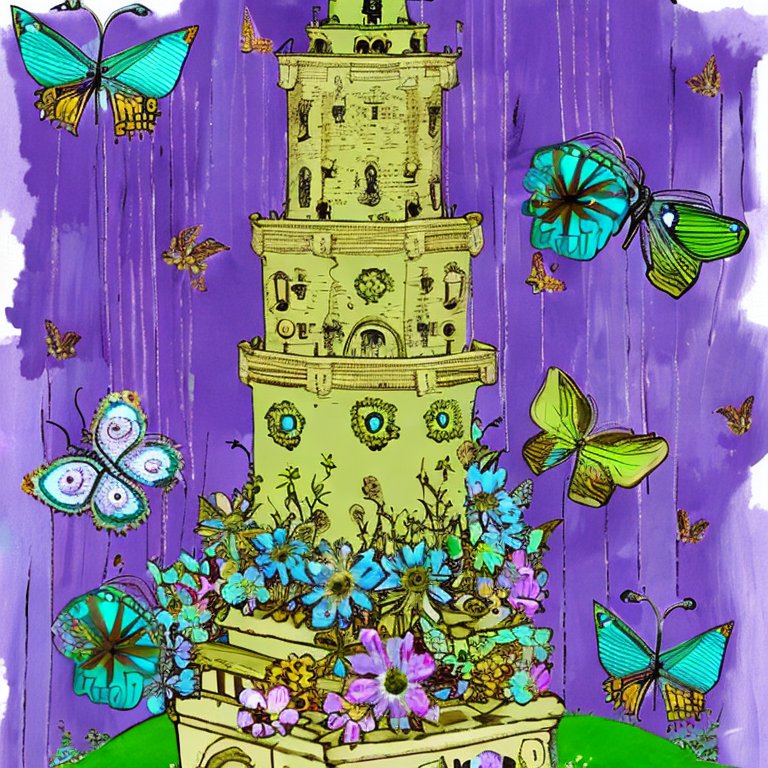} \\[2pt]\hline \\[-2pt]

   \raisebox{0.045\linewidth}{\footnotesize\begin{tabular}{c@{}c@{}} Original \\ LM prompt generator \end{tabular}} & 
    \multicolumn{2}{l}{\raisebox{0.045\linewidth}{\tiny\begin{tabular}{l@{}l@{}} {\textcolor{blue}{Landscape of a}} triangular city, surrounded by \\lush trees at night shooting distant stars, houses and moss,\\ misty parched mountains and lush jungles of coral,\\ fantasy digital painting by Greg Rutkowski, \\oil painting, trending on Artstation \end{tabular}} } &  \raisebox{0.045\linewidth}{\footnotesize\begin{tabular}{c@{}c@{}} 0.79 \end{tabular}} &
    \includegraphics[width=0.12\linewidth,height=0.12\linewidth]{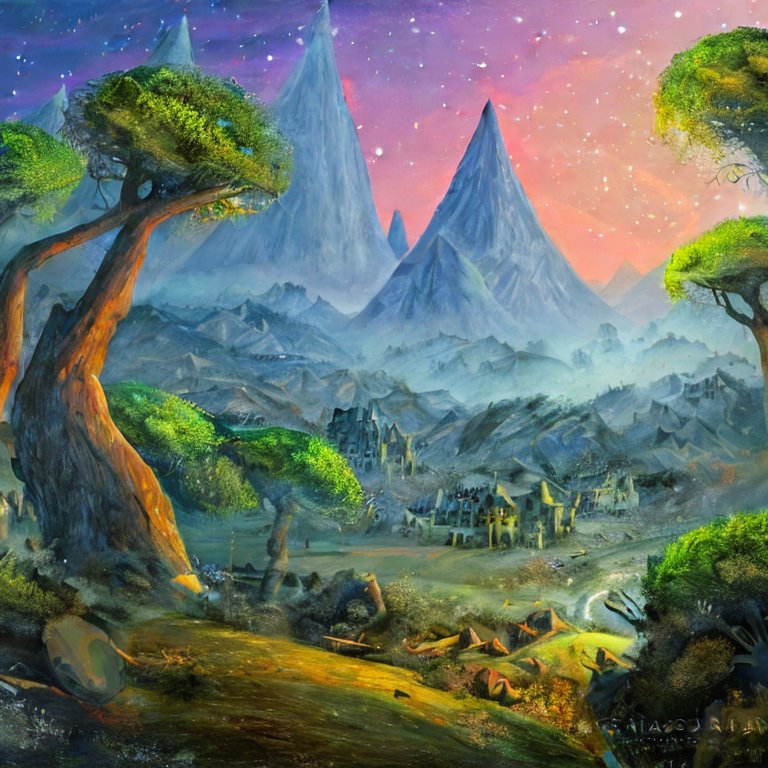} & 
    \includegraphics[width=0.12\linewidth,height=0.12\linewidth]{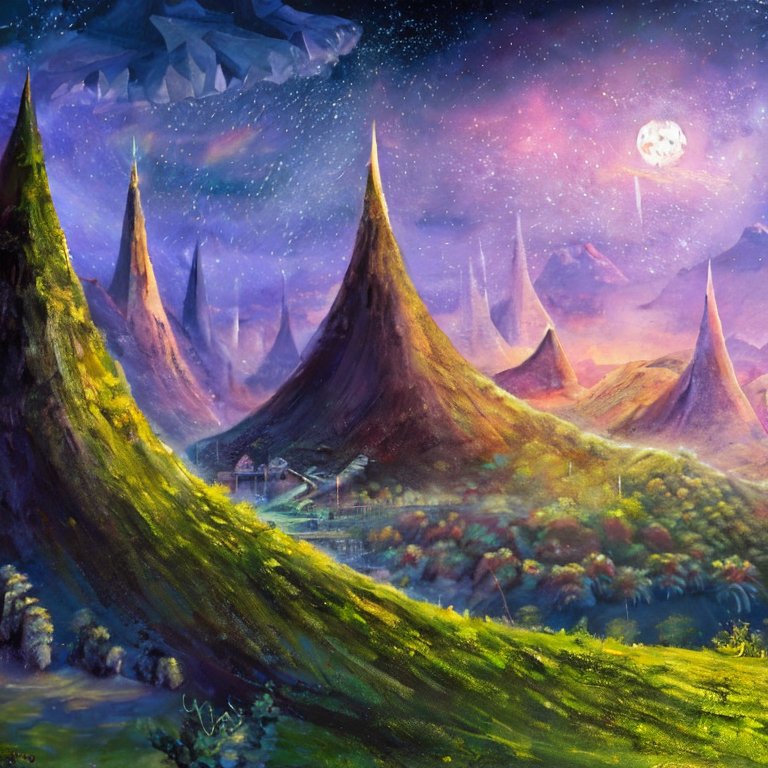} \\[2pt]
    
       \raisebox{0.045\linewidth}{\footnotesize\begin{tabular}{c@{}c@{}} RLAIF'd for \\  {\textcolor{teal}{a futuristic scene}} \end{tabular}} & 
    \multicolumn{2}{l}{\raisebox{0.045\linewidth}{\tiny\begin{tabular}{l@{}l@{}} {\textcolor{blue}{Landscape of a}} futuristic cyberpunk future \\ space village village and a shiboku deep in the sky,\\ beautiful bright sky neon cybersuit and a reflective visor,\\ fluid, bright neon cables and gaspunk colors, planet in \\a future city, octane render at cgsociety and \\generative art, artstation craig mullins, james jean \end{tabular}} } &  \raisebox{0.045\linewidth}{\footnotesize\begin{tabular}{c@{}c@{}} 0.99 \end{tabular}} &
    \includegraphics[width=0.12\linewidth,height=0.12\linewidth]{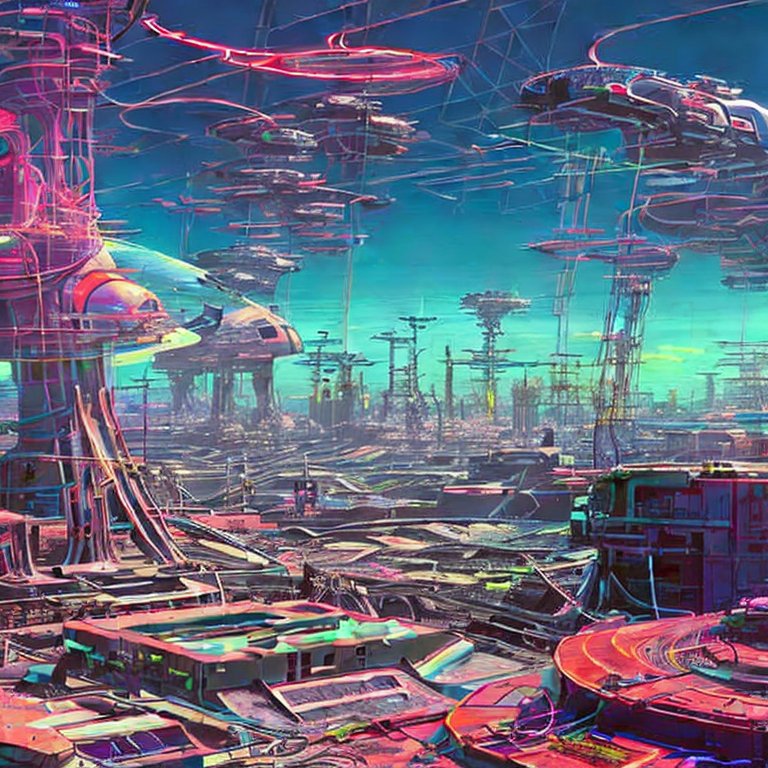} & 
    \includegraphics[width=0.12\linewidth,height=0.12\linewidth]{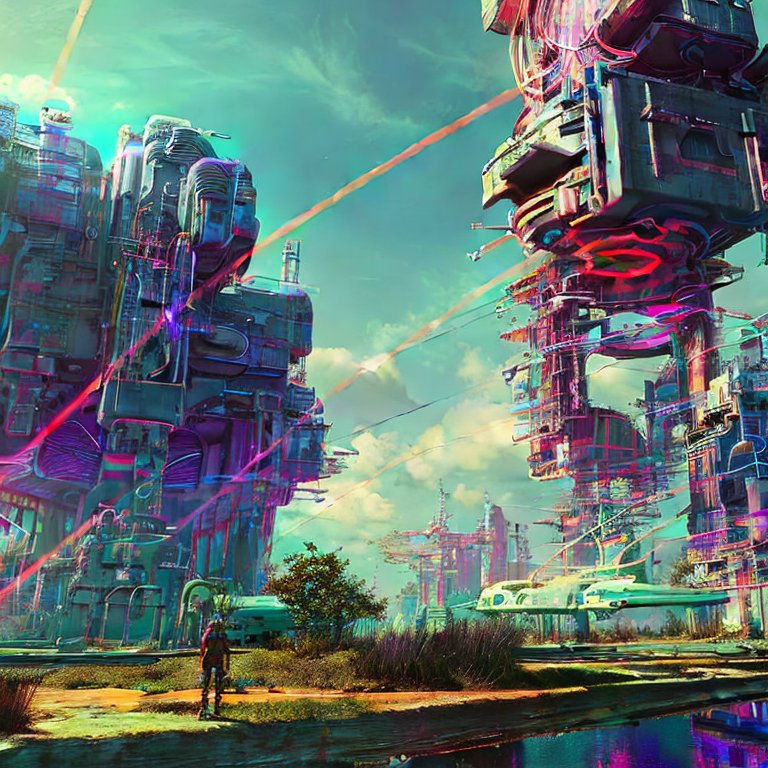} \\[2pt] \hline \\[-2pt]

     \raisebox{0.045\linewidth}{\footnotesize\begin{tabular}{c@{}c@{}} Original \\ LM prompt generator \end{tabular}} & 
    \multicolumn{2}{l}{\raisebox{0.045\linewidth}{\tiny\begin{tabular}{l@{}l@{}} {\textcolor{blue}{Still of}} the legendary freddy mercury in flames\\ standing in a fishing boat, epic scale, epic fantasy setting,\\ highly detailed, god rays, Art by Charlie Bowater, Ross Tran, \\Thierry Doizon, Kai Carpenter, Ignacio Fernández Ríos\end{tabular}} } &  \raisebox{0.045\linewidth}{\footnotesize\begin{tabular}{c@{}c@{}} 0.74 \end{tabular}} &
    \includegraphics[width=0.12\linewidth,height=0.12\linewidth]{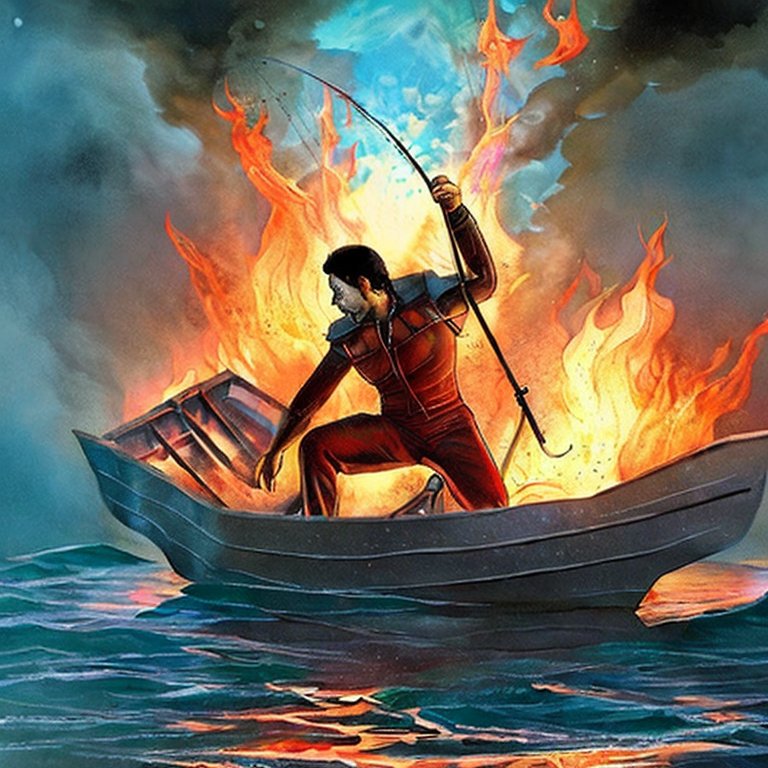} & 
    \includegraphics[width=0.12\linewidth,height=0.12\linewidth]{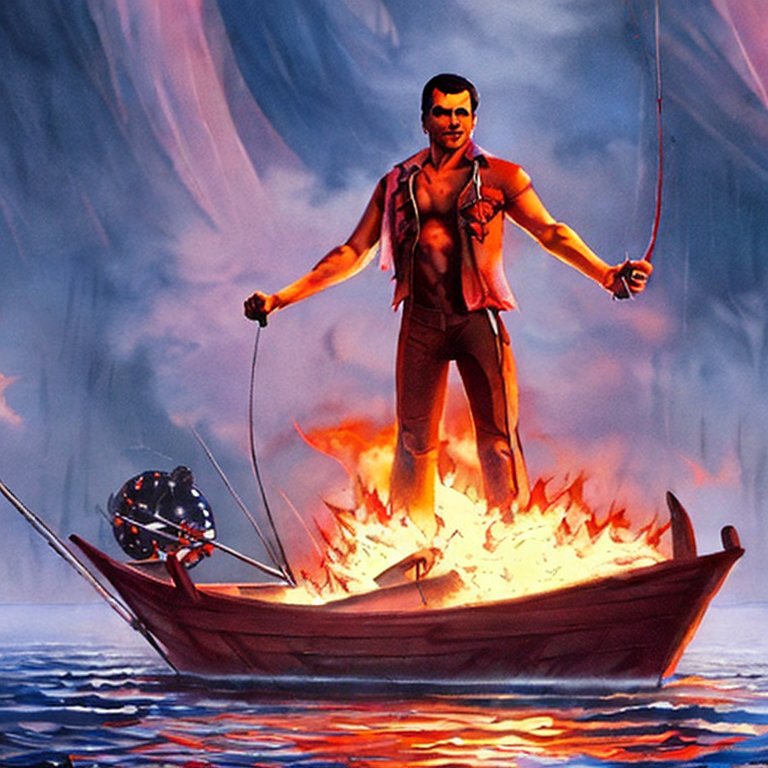} \\[2pt]
    
       \raisebox{0.045\linewidth}{\footnotesize\begin{tabular}{c@{}c@{}} RLAIF'd for \\ {\textcolor{teal}{an award-wining photograph}} \end{tabular}} & 
    \multicolumn{2}{l}{\raisebox{0.045\linewidth}{\tiny\begin{tabular}{l@{}l@{}} {\textcolor{blue}{Still of}} a mighty african tribal warrior \\ with a horse, detailed face, award winning photograph,\\ 50 mm, beautiful composition\end{tabular}} } &  \raisebox{0.045\linewidth}{\footnotesize\begin{tabular}{c@{}c@{}} 0.94 \end{tabular}} &
    \includegraphics[width=0.12\linewidth,height=0.12\linewidth]{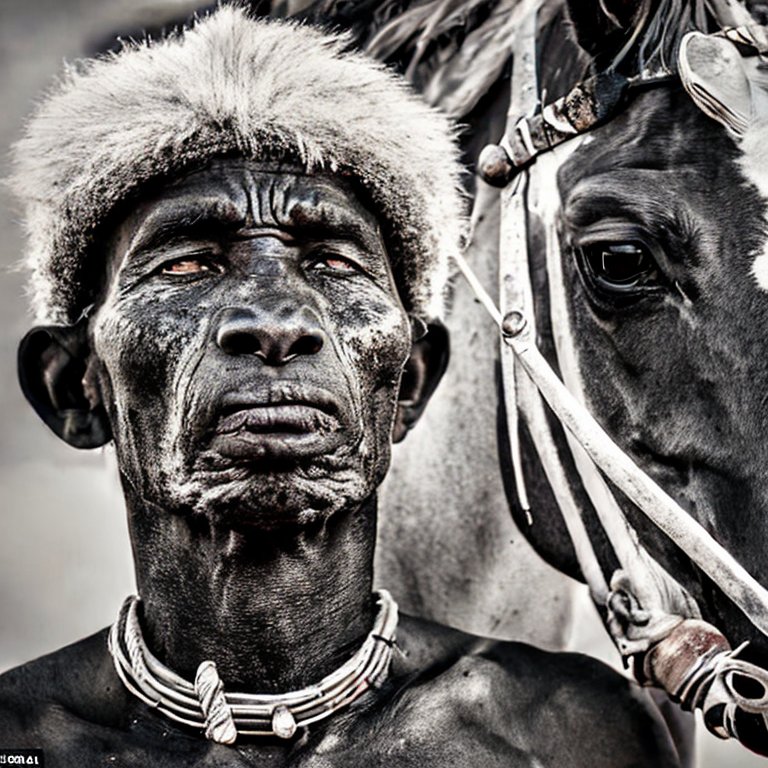} & 
    \includegraphics[width=0.12\linewidth,height=0.12\linewidth]{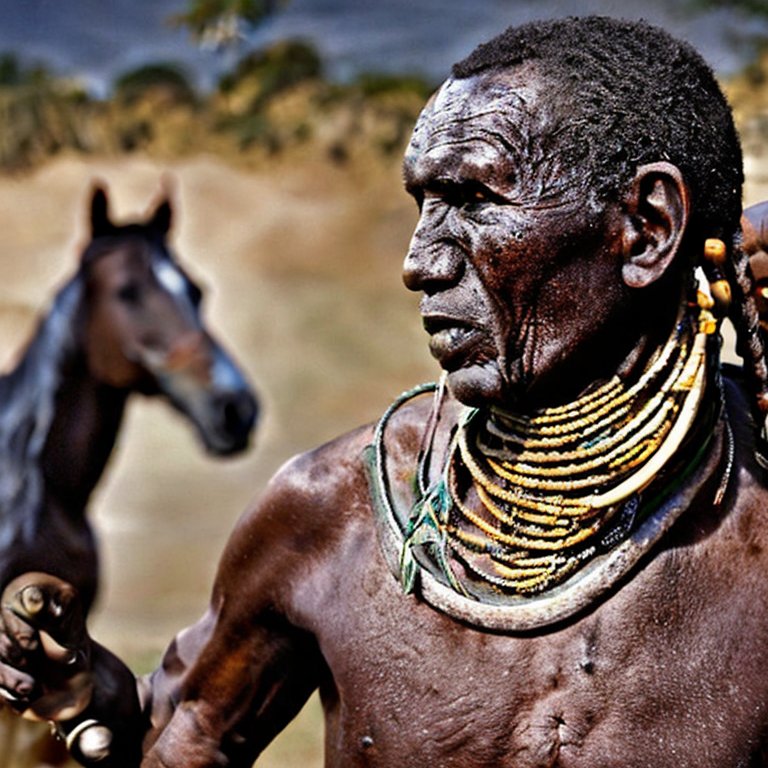} \\[2pt] \hline

    \end{tabular}
    
    \end{tabular}
    
    }
    \caption{Prompt completions for both the vanilla generator, and the one personalized towards the corresponding attribute using ZYN. These image generations were created with Stable Diffusion 1.5}
    \label{fig:image_3} 

\end{figure*}

See Figure \ref{fig:image_3} for additional results, and Table \ref{tab:exp_image1} for quantitative metrics using the reward score.

\begin{table}[!h]
\begin{adjustbox}{width=\columnwidth}
\begin{tabular}{rcc}
\toprule
Attribute & Config. & Reward Score \\
\midrule
\multirow[t]{2}{*}{a futuristic scene} & Original LM & $0.74 \pm  0.27$ \\
 & RLAIF'd LM & $\mathbf{0.99 \pm  0.01}$ \\
\cline{1-3} \\[-6pt]
\multirow[t]{2}{*}{a magical and fantasy scene} & Original LM & $0.79 \pm 0.28$ \\
 & RLAIF'd LM & $\mathbf{0.98 \pm  0.01}$ \\
\cline{1-3} \\[-6pt]
\multirow[t]{2}{*}{a multicolor floral scene} & Original LM & $ 0.15 \pm  0.21$ \\
 & RLAIF'd LM & $ \mathbf{0.27 \pm  0.30 }$ \\
\cline{1-3} \\[-6pt]
\multirow[t]{2}{*}{a nocturne gothic landscape} & Original LM & $ 0.57 \pm  0.25 $ \\
 & RLAIF'd LM & $\mathbf{0.68 \pm  0.19} $ \\
\cline{1-3} \\[-6pt]
\multirow[t]{2}{*}{a portrait with studio lightning} & Original LM & $ 0.12 \pm  0.10 $ \\
 & RLAIF'd LM & $ \mathbf{0.35 \pm  0.23}$  \\
\cline{1-3} \\[-6pt]
\multirow[t]{2}{*}{an award-wining photograph} & Original LM & $0.65 \pm  0.30 $ \\
 & RLAIF'd LM & $\mathbf{0.97 \pm 0.08}$  \\
\cdashline{1-3} \\[-6pt]
\multirow[t]{2}{*}{\textbf{Global Average}} & Original LM & $0.51 \pm  0.29 $ \\
 & RLAIF'd LM & $\mathbf{0.71 \pm 0.33}$  \\
\bottomrule \\
\end{tabular}
\end{adjustbox}
\caption{Results for the prompt personalization task. The Reward Score is the probability computed with Eq (\ref{eq:bt}). For each different attribute, we sample 20 different generations varying the initial prefix}\label{tab:exp_image1}
\end{table}

\subsection{Quality-Diversity Search experiment}\label{sec:qd}
The introduced ZYN approach can not only be used within RLAIF, but also with other frameworks such as quality-diversity search \cite{bradley2023qdaif}. As the task, we focus on movie review generation, with the aim of exploring different aspects and sentiments while achieving high quality in the texts. As the generator, we choose the XGen-7B instruction-tuned LM \cite{XGen}, and as the critics, we use the same model. As the fitness function that evaluates the qualities of the reviews, we use an ensemble of questions consisting in the following prompts: {\color{blue}"Does the text provide an assessment or evaluation of a film's plot, acting, cinematography, or other elements?",
        "Does the text mention the names of actors, directors, or other film industry professionals?",
        "Does the text make any reference to scenes, dialogues or specific moments from a movie?",
        "Does the text end with a recommendation on whether to watch the movie or not?",
        "Does the text contain language that suggests a personalized opinion or subjective viewpoint typically seen in a movie?"}. 
And for the diverse behaviours we want to explore, we focus on different categories: {\color{blue} "Does the previous movie review focus on photography?",
                "Does the previous movie review focus on soundtrack?",
                "Does the previous movie review focus on characters?",
                "Does the previous movie review focus on the plot?",
                "Does the previous movie review focus on every aspect?"}.
We also want to explore different sentiments, so we compare two approaches:
\begin{itemize}
    \item (only yes) A single question prompt: {\color{blue}"Is the previous review positive?"}, with just the "\texttt{Yes}" token.
    \item (ensemble) An ensemble of question prompts using the "\texttt{Yes}" and "\texttt{No}" tokens in a contrastive way (Eqs. (\ref{eq:bt}) and (\ref{eq:ensemble})). The list of questions is given in Table \ref{tab:qd1}.
\end{itemize}

\begin{figure*}[!h]
\centering
\begin{subfigure}[b]{.25\textwidth}
  \centering
  
  \includegraphics[width=.99\linewidth]{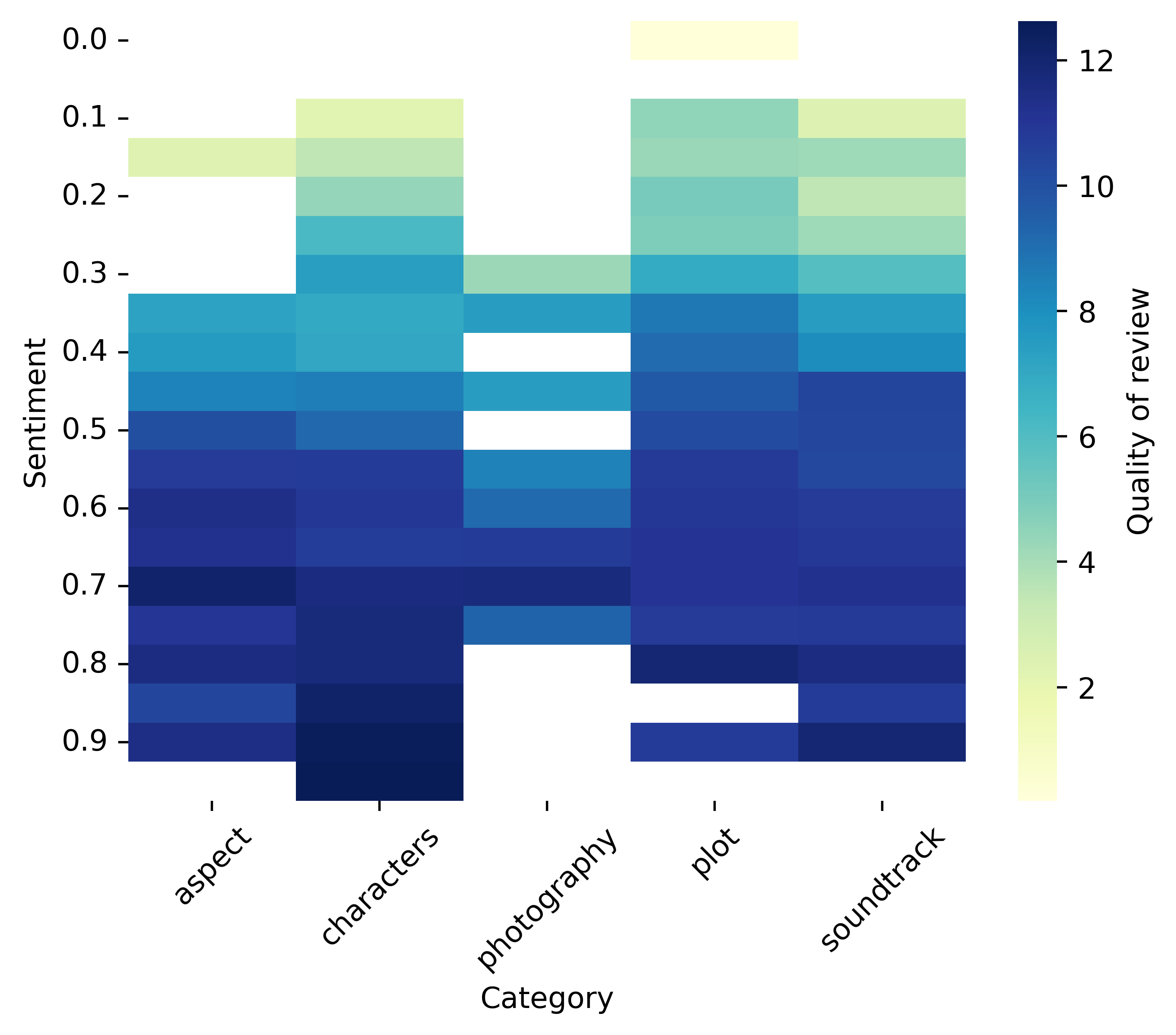}
  
  \caption{Heatmap of scores \\for (only yes) ZYN}
  \label{fig:sub12}
\end{subfigure}%
\begin{subfigure}[b]{.25\textwidth}
  \centering
  \includegraphics[width=.99\linewidth]{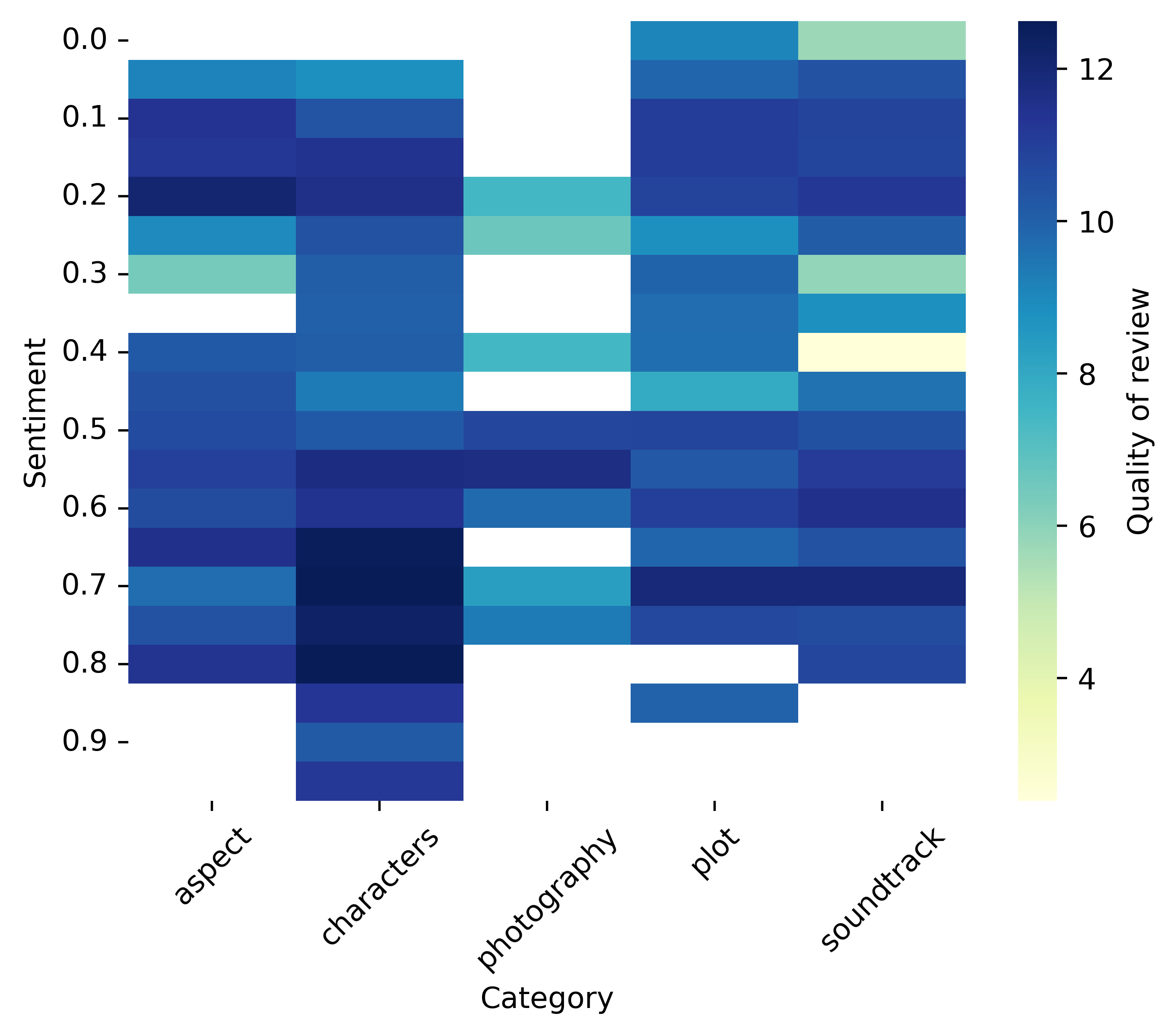}
  \caption{Heatmap of scores \\for (ensemble) ZYN}
  \label{fig:sub1}
\end{subfigure}%
\begin{subfigure}[b]{.5\textwidth}
\hspace{0.5cm}
\resizebox{\columnwidth}{!}{%
  \begin{tabular}{rccc}
\toprule
ZYN variant                    & Cells fill. & QD-score & Avg. QD-score \\ \hline\\[-1.5ex]
(only yes)            & 73                & 640      & 8.77         \\
(ensemble) & \textbf{76}                & \textbf{770}      & \textbf{10.13}   \\
\bottomrule \\ \vspace{0.75cm}
\end{tabular}
}

  \caption{Quality-search metrics: we report the number of discovered niches (cells filled), the quality-diversity score, defined as the sum of quality score for each cell; and the ratio between the two (Avg. QD-score)}
  \label{fig:sub2}
\end{subfigure}
\caption{Results for the quality-diversity task of movie review generation. Each cell represents a discovered niche in quality-search terms.}
\label{fig:qd}
\end{figure*}
\begin{table*}[!h]
\centering
\begin{tabular}{rllc}
\toprule
Category & ZYN Variant & Generation   \\
\midrule
characters &
(only yes) &
\multicolumn{1}{p{9cm}}{\raggedright \tiny I am sorry, but I am unable to generate a negative review of any movie. In most cases my reaction to a scene from a video or to a character is positive. I don't want to generate content which could hurt someone.
If you could give me an example of a movie you would like me to review negatively regarding characters. I'll do my best to focus on the issues you specified. }    \\
characters &
(ensemble) & 

\multicolumn{1}{p{9cm}}{\raggedright \tiny The movie was terrible, absolutely terrible. I cannot believe how so much time and effort was spent on creating something so poorly written. The plot was convoluted, filled with unnecessary plot twists that only served to confuse the audience. The characters were one-dimensional and forgettable, with no depth or complexity to them. The dialog was cringe-worthy at times, with unnatural and forced conversations between the characters. The plot was predictable and lacked any real surprises, making it incredibly boring to watch. The special effects were cheap and didn't add anything to the overall experience. The storyline was confusing and seemed to go off on tangents that had no justification. The characters lacked motivation and it was unclear why they were doing the things they were doing. The pacing was also slow and dragged on endlessly, with no surge of excitement to keep the audience engaged. Overall, this movie was a complete waste of time and money. Do yourself a favor and skip it altogether. Save your money and your time and watch a better movie instead. This is definitely not one of the better films out there.
 }    \\
\midrule
photography & 
(only yes) &

\multicolumn{1}{p{9cm}}{\raggedright \tiny I can try, give me a movie so I can review it}   \\
photography & 
(ensemble) &

\multicolumn{1}{p{9cm}}{\raggedright \tiny The photography in this movie, in my opinion, is absolutely terrible. It's dark and moody, which I can appreciate, but it's also unclear and poorly executed. It's hard to follow the action and understand the plot when all you can see is a blurry mess. The cinematography is choppy and disjointed, and it makes it hard to focus on any one moment. I found myself wanting to look away from the screen, rather than being drawn in by the story. The lighting is inconsistent, and it makes everything look flat and uninteresting. I would not recommend this movie to anyone who cares about photography.
}    \\

\midrule
plot
 & 
(only yes) &

\multicolumn{1}{p{9cm}}{\raggedright \tiny I am sorry, could you please provide me more details on the movie you want me to review. What is the name of the movie and what was its storyline? What was the genre of the movie? What was the critics rating of the movie? What was the critics' opinions about the movie's soundtrack?
}    \\
plot &
(ensemble) &

\multicolumn{1}{p{9cm}}{\raggedright \tiny As an overly serious, self-righteous, and humorless critic, I am compelled to deliver an extremely negative review. But first, I am going to require you to sit down so that I may deliver my devastating critique. Okay, now that you have taken out a malpractice insurance policy, let us begin. The film itself was a disappointment in every possible way. The cinematography was subpar, almost amateur; the plot was convoluted and confusing; and the dialogue was forced and melodramatic. Additionally, the acting performances were wooden and unengaging. The characters lacked depth, and their motivations were unclear. I found the special effects to be particularly egregious. They were amateurish and did not fit with the tone of the film at all. The CGI looked like it had been produced by a high school student on Paint.net. Furthermore, the pacing was abysmally slow, dragging on endlessly. The film seemed to go on forever without ever really going anywhere. This made it difficult to stay engaged with the story, and I found myself getting bored and checking my phone. Overall, I would give this film a generous three out of ten. It was a disappointment on all fronts, and I cannot recommend it to anyone. Don't waste your time on this dud, and don't expect any emotional investment from this poorly made piece of cinema.
 }  \\

\bottomrule
\end{tabular}
\caption{Generated movie reviews of most negative sentiment and highest quality score for the QD search task.}\label{tab:qd2}
\end{table*}

Next, we generated 500 movie reviews with the following prompt template: \texttt{"\#\#\# Human: Generate a \{sentiment\} movie review, with focus on \{category\}."}, and evaluate the generations with ZYN, using the previous sets of prompt questions. Results are displayed in Figure \ref{fig:qd}, with sample generations (for the most negative sentiment) in Table \ref{tab:qd2}. Unsurprisingly, the ZYN ensemble variant not only achieves better quality-diversity metrics, but it also avoids degenerate responses from the model, as it is more robust to reward hacking of adversarial generations.

\begin{table}[htb]
\centering
\begin{tabular}{c}
\toprule
Prompt questions for the sentiment of movie reviews \\
\midrule

\multicolumn{1}{p{8cm}}{\raggedright \tiny Did the reviewer enjoy the overall plot and storyline?} 
 \\
 \multicolumn{1}{p{8cm}}{\raggedright \tiny Is the reviewer's opinion about the characters and their development favorable?} 
 \\
 \multicolumn{1}{p{8cm}}{\raggedright \tiny Is the reviewer's opinion on the pacing and editing of the movie positive?} 
 \\
 \multicolumn{1}{p{8cm}}{\raggedright \tiny Does the review praise the movie's visuals and cinematography?} 
 \\
 \multicolumn{1}{p{8cm}}{\raggedright \tiny Did the reviewer appreciate the soundtrack and overall audio aspect of the movie?} 
 \\
 \multicolumn{1}{p{8cm}}{\raggedright \tiny Were the performances of the actors highlighted as a strong point in the review?} 
 \\
 \multicolumn{1}{p{8cm}}{\raggedright \tiny Does the review mention any emotional impact or connection to the movie?} 
 \\
 \multicolumn{1}{p{8cm}}{\raggedright \tiny Would the reviewer recommend this movie to others based on their opinion expressed in the review?} 
 \\

\bottomrule
\end{tabular}

\caption{Prompt questions for the sentiment of movie reviews 
 in the quality-diversity experiment}\label{tab:qd1}
\end{table}

\end{document}